%% file: main.tex
\documentclass[10pt,twocolumn,letterpaper]{article}

\usepackage{wacv}              % To produce the CAMERA-READY version

\definecolor{wacvblue}{rgb}{0.21,0.49,0.74}
\usepackage[pagebackref,breaklinks,colorlinks,allcolors=wacvblue]{hyperref}

\def\wacvPaperID{16} % *** Enter the WACV Paper ID here
\def\confName{WACV}
\def\confYear{2026}
\usepackage{float}
\usepackage[ruled]{algorithm2e} % For algorithms

\usepackage{amssymb}
\usepackage{graphicx,dblfloatfix}
\usepackage{natbib}
\usepackage{amsmath}
\usepackage{eccvabbrv}
\usepackage{booktabs}
\usepackage{multirow}
\usepackage{footnote}
\usepackage{caption}
\usepackage{subcaption}
\usepackage{xcolor}
\usepackage{soul}
\usepackage{hyperref}
\usepackage[capitalize]{cleveref}

\newcommand{\myparagraph}[1]{\vspace{0pt}\noindent{\bf #1}}

\begin{document}

\title{CraftSVG: Multi-Object Text-to-SVG Synthesis via Layout Guided  Diffusion}

\author{Ayan Banerjee$^1$, Nityanand Mathur$^2$, Josep Lladós$^1$, Umapada Pal$^3$, Anjan Dutta$^4$ \\
$^1$\textit{Computer Vision Center, Universitat Autònoma de Barcelona (\{abanerjee,josep\}@cvc.uab.es)}, \\
$^2$\textit{Smallest AI (nityanand@smallest.ai);} $^3$\textit{Indian Statistical Institute, Kolkata (umapada@isical.ac.in)} \\ $^4$\textit{Institute for People Centred Artificial Intelligence, University of Surrey (anjan.dutta@surrey.ac.uk)}\\}

\maketitle

\input{tex/00_abs}

\input{tex/01_intro}

\input{tex/02_related_backup}

\input{tex/03_method}

\input{tex/04_expts}

\input{tex/05_concl}

\bibliographystyle{elsarticle-num}
\bibliography{biblio}

\clearpage

\input{tex/06_supply}

\end{document}

%% file: tex/00_abs.tex
\begin{abstract}
Generating SVGs from text is a challenging vision task, requiring diverse yet realistic depictions of the seen as well as unseen entities. Existing research has been mostly limited to generating single-object rather than comprehensive scenes comprising multiple elements. In response, CraftSVG introduces an end-to-end framework for creating SVGs depicting entire scenes from a textual description. Utilizing a pre-trained LLM for layout generation from text via iterative in-context learning, CraftSVG introduces a per-box mask latent mechanism for accurate object placement. A fusion mechanism is developed to integrate the attention maps, employing a diffusion U-Net for coherent composition, which accelerates stroke initialization. Recognizing the importance of abstract SVGs in communication, we incorporated an MLP-based mechanism to simplify SVGs, with alignment and perceptual loss via differential rendering and opacity modulation to improve aesthetics. CraftSVG outperforms previous methods in abstraction, recognizability, and detail, as depicted by its CLIP-T: 0.5013,
Aesthetic: 7.0779, score. The code is available at \href{https://github.com/ayanban011/SVGCraft}{github.com/CraftSVG}.
\vspace{-10pt}
\end{abstract}

%% file: tex/01_intro.tex
\section{Introduction}
\label{sec:intro}

\begin{figure}[t]
%\vspace{-10.5pt}
  \centering
   \includegraphics[width=\columnwidth]{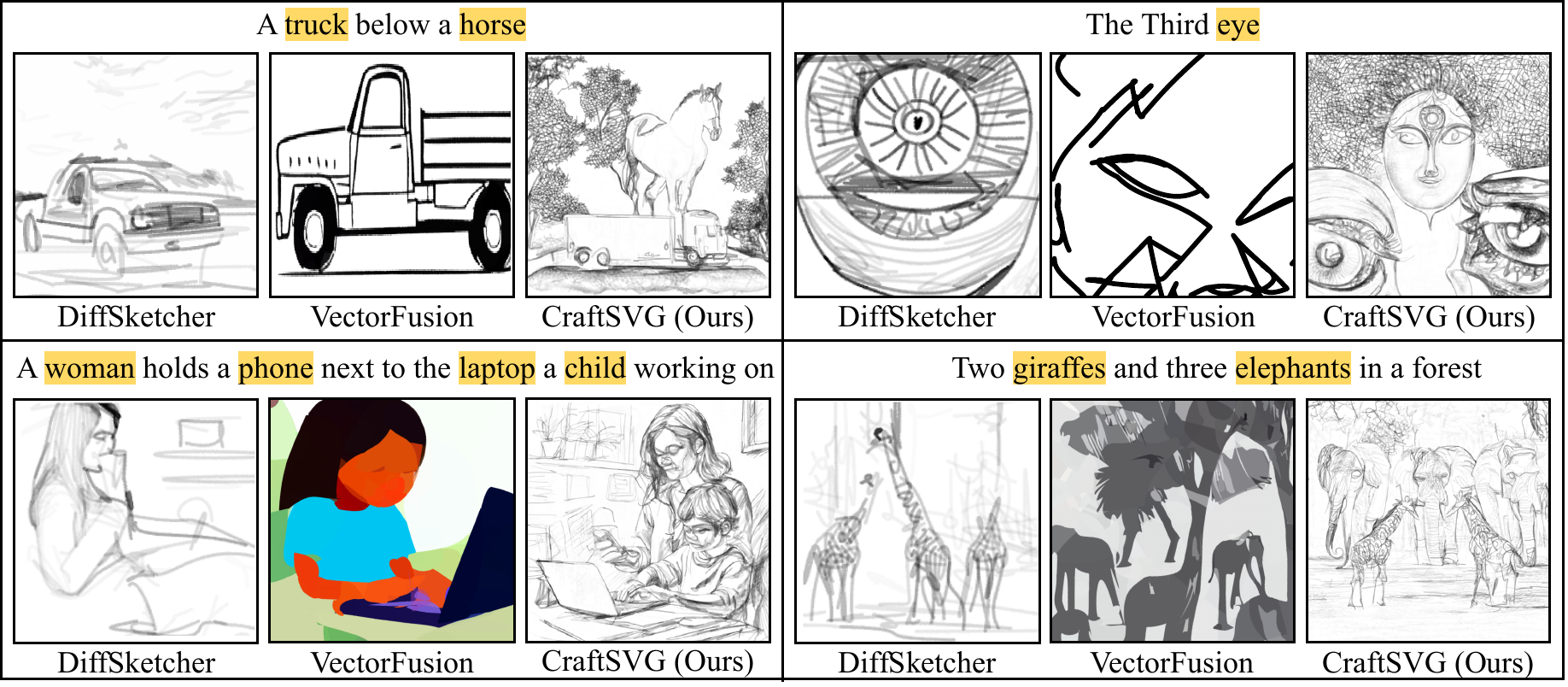}
   
    \caption{\textbf{Previous methods \cite{xing2023diffsketcher,jain2023vectorfusion} vs Ours}: While former methods are unable to generate the accurate numeration, spatial relationship, and imaginary concept, CraftSVG ensures all through the layout guidance. (Note:  \#strokes = 1024; stroke width = 4.0).}
    \label{fig:intro}
    \vspace{-6mm}
\end{figure}

Graphic representations, including Scalable Vector Graphics (SVG), are crucial in enhancing communication by transforming abstract concepts and complex information into detailed visuals, thereby improving comprehension. Utilizing basic elements like B\'{e}zier curves and polygons, SVGs are integral in digital design for their scalability across devices with minimal file sizes. Research efforts like CLIPDraw \cite{frans2022clipdraw} leverage CLIP text and image encoders \cite{radford2021learning} for text-based image creation. In contrast, advancements like VectorFusion \cite{jain2023vectorfusion} and DiffSketcher \cite{xing2023diffsketcher} employ diffusion models \cite{Rombach2022LatentDiffusion} to refine SVGs, enhancing the quality and fidelity of SVG outputs. However, these models struggle with complex prompts involving relationships, enumeration, spatial arrangements, and imaginative concepts, due to limitations in their prompt encoding techniques, which focus on single objects or a global perspective (see \cref{fig:intro}). They also utilized Score Distillation Sampling (SDS) loss, which often produces smooth and detail-less results, and the models typically suffer from a lack of diversity. Although developing a comprehensive multi-modal dataset can address these issues, it requires extensive resources, and may not capture imaginative concepts beyond real-world objects.

To address the aforementioned challenges, we propose CraftSVG, an end-to-end method that utilizes a pre-trained large language model (LLM) to generate layouts from text prompts via iterative in-context learning \cite{wu2023self}. Given a text prompt, the LLM is adept at producing canvas layouts as captioned bounding boxes. Subsequently, we implement a per-box mask latent-based canvas initialization method that eliminates unnecessary strokes without post-processing, ensuring quicker convergence with more precise control than bounding boxes by providing the shape/structure of the elements. Recognizing the importance of abstract SVGs in communication, we developed a mechanism to abstractify SVGs at different levels.  This involves training two multilayer perceptrons (MLP) in parallel: the first ($\text{MLP}_s$) optimizes the placement of foreground object strokes, while the second ($\text{MLP}_d$) gradually removes background strokes without compromising scene recognizability or semantics. To ensure proper alignment of foreground and background strokes, we propose a perceptual alignment loss, combined with semantic-aware opacity modulation, to maximize perceptual similarity for output SVG generation.

In summary, our contributions are as follows: (1) An end-to-end architecture for SVG synthesis via \textit{stroke optimization} from complex textual prompts that describe relationships, \textit{enumeration}, and \textit{spatial arrangements} of \textbf{\textit{multiple objects}}; (2) A method for canvas initialization using a \textbf{\textit{per-box mask latent}} followed by \textit{iterative} layout correction with \textbf{\textit{in-context learning}}, which effectively eliminates unnecessary strokes without post-processing and leads to quicker convergence; (3) An MLP based SVG abstraction with \textbf{\textit{perceptual alignment loss}} to ensure perfect stroke alignment between foreground and background with \textbf{\textit{semantic aware opacity modulation}} to produce the depth effect for providing the sketches an artistic look; (4) Extensive experiments to confirm the effectiveness of CraftSVG, which outperforms existing methods by an effective margin.

\begin{figure*}[!htbp]
\centering
  \includegraphics[width=\textwidth]{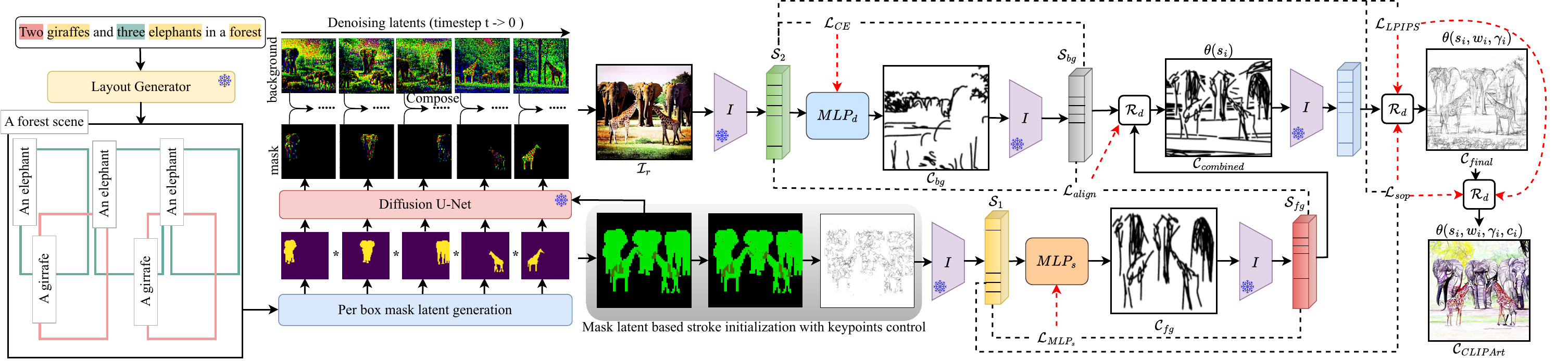}
\caption{\textbf{CraftSVG} employs an LLM to generate layouts with a ``background prompt'', ``grounding object'' and their corresponding bounding boxes. Masked latents for each box, with controlled attention, ensure accurate object placement. These latents are fused to initialize the SVG canvas, which is used by a diffusion U-Net for coherent image generation ($\mathcal{I}_r$) that aligns with the layout. The final canvas is produced via two parallel-trained MLPs using perceptual alignment loss and opacity modulation, maximizing the similarity between $\mathcal{I}_r$ and $\mathcal{C}_\text{CLIPArt}$.}
\label{fig:sketchcraft}
\vspace{-4mm}
\end{figure*}

%% file: tex/02_related_backup.tex
\section{Related Work}
\label{sec:sota}
\myparagraph{SVG Generation:}
Early SVG generation focused on sketch generation, starting with edge map extraction \cite{xie2015holistically,li2019photo,liu2020unsupervised,chen2020deepfacedrawing,tong2021sketch} and advancing to higher abstractions via seq-to-seq models \cite{li2020sketch,lin2020sketch,das2022sketchode,das2021cloud2curve} optimizing SVG strokes. These methods generated sketches specific to training data, limiting out-of-distribution SVG generation. Differential rendering \cite{li2020differentiable} expanded SVG capabilities, with CLIP \cite{radford2021learning} via advanced visual text embedding, leading to CLIPDraw \cite{frans2022clipdraw}, which aligns image and text embeddings using gradient descent on Bézier curves. However, CLIPDraw often produces messy SVGs and struggles with unseen concepts. StyleCLIPDraw \cite{schaldenbrand2022styleclipdraw} incorporated human-defined styles but didn't reduce noise or preserve textual intent. CLIPasso \cite{vinker2022clipasso} and CLIPascene \cite{vinker2023clipascene} generate image-conditioned sketches via CLIP embeddings but struggle with complex scenes due to CLIP's attention limitations. CraftSVG addresses this limitation with a per-box mask-latent based initialization, enabling multi-object SVG generation with accurate spatial relationships.

\myparagraph{Diffusion Models:} 
The advent of Probabilistic Models (DDIMs, DDPMs) \cite{wang2022sketchknitter,das2023chirodiff} has advanced SVG synthesis, yet complex freehand sketches remained challenging until T2I diffusion models \cite{pmlr-v162-nichol22a,rombach2022high,ramesh2022hierarchical,saharia2022photorealistic} showed promising results in synthesizing images from text. DiffSketcher \cite{xing2023diffsketcher} and VectorFusion \cite{jain2023vectorfusion} both employed diffusion models for optimizing parametric SVGs. However, their focus on single objects, due to initialization from U-Net attention, motivated the development of CraftSVG, which introduces per-box mask latent-based initialization for comprehensive scene generation with multiple objects. Unlike SVGDreamer \cite{xing2023svgdreamer}, which prioritizes style over spatial relationships in SVG synthesis, CraftSVG excels in depicting complex scenes, setting it apart from other SVG synthesis methods.

\myparagraph{Multi Object Generation:}
To our knowledge, no prior method supports multi-object text-to-SVG generation. Some T2I generation methods \cite{lian2023llm,ge2023expressive,wu2023self,qu2023layoutllm,zhou2024migc,chen2024training,ohanyan2024zero,qu2023layoutllm} address multi-object image generation but suffer from scalability issues and fail to maintain object interactivity with the background (\cref{fig:comp_diff} in the Appendix). CraftSVG ensures object interactivity through alignment loss, stroke initialization in the foreground region, and optimization of SVG parameters, enabling artistic and realistic SVG generation.

%% file: tex/03_method.tex
\section{CraftSVG}
\label{sec:method}

\subsection{Problem formulation}
\label{sec:3.0}
The SVGs are defined as a set of B\'{e}zier curves $ \theta = \{S_i\}_{i=1}^n = \{s_i,w_i, \gamma_i, c_i \}_{i=1}^n$ where $s_i$ represents, its strokes attribute with $p$ control points such that $s_i = \{x_{i_j}, y_{i_j}\}_{j=1}^p$ for any $p\in[1,\infty)$ along with stroke width $w_i$ and opacity $\gamma_i$. Here, $c_i$ represents the corresponding color attributes of the B\'{e}zier curve $S_i$ such that $c_i = (r, g, b, a)_i$. Our optimization-based SVG generation process starts with initializing a canvas $\mathcal{C}$ with a set of B\'{e}zier curves $\{S_i\}_{i=1}^n$ and then optimize $\mathcal{C}$ by back-propagating the gradient of rasterized image $\mathcal{I}_r$ to SVG parameters $\theta$ via a differential renderer $\mathcal{R}_d(\theta)$. This optimization controls the amount of projective transformations (8 dof: rotations, translations, anisotropic-scaling, and morphing; each poses 2 dof) that $\theta$ undergoes to obtain the maximum similarity with $\mathcal{I}_r$.

%Primitive-based canvas completion performs the same synthesis through optimization in a more constrained environment. A primitive shape is a special form of B\'{e}zier curves $\{S_i\}_{i=1}^n = \{s_i,w_i, \gamma_i \}_{i=1}^n$, where $S_i = \{s_i\}_{j=1}^p = \{x_i, y_i\}_{j=1}^p; \forall p\in\{2,3,4\}$ and they can only go through the affine transformation (6 dof: rotation, translation and anisotropic-scaling) during optimization with $\mathcal{R}_d$. This effect of transformation constraints has been provided in \cref{fig:exp1}, \cref{tab:02} and \cref{fig:ab_shape_evol}, \cref{fig:ab_shape_evol_two}, in the appendix.

CraftSVG utilizes a T2I diffusion model as a prior to direct the differentiable renderer $\mathcal{R}_d$ and optimizes the parametric graphics $\theta$ leading to the SVG synthesis that aligns with the provided textual description $[T_p]$ (see \cref{algo1}). This end-to-end pipeline consists of four stages(see \cref{fig:sketchcraft}). 
We obtain a layout from the provided text prompt $[T_p]$ using LLM with iterative in-context learning to allow precise region control over text embeddings. We designed a mechanism called ``per box mask latent generation'' which guides the diffusion U-Net to generate object instances in designated regions and helps in canvas initialization by providing the control points of the B\'{e}zier curves through the proposed fusion mechanism. Finally, we optimize the canvas by maximizing similarity between the target image $\mathcal{I}_m$ and the canvas $\mathcal{C}$ via perceptual loss.

%Given a text prompt, CraftSVG creates abstract-to-dense SVGs following the enumeration and spatial relationship as depicted in \cref{fig:sketchcraft}. Here, we first generate a layout from text via LLM with poroposed similarity based iterative correction, which allows precise region control over the textual description. We utilize a per-box mask latent-based attention mechanism that helps in canvas initialization and guides the target SVG generation with an MLP-based abstraction via perceptual alignment loss. Then, we propose a semantic-aware opacity optimization to provide shading and depth effect, and we jointly use this opacity control along with perceptual loss to optimize the SVGs.
%\vspace{-6mm}
\begin{algorithm}[!htbp]
\caption{CraftSVG: Text2SVG synthesis}
\SetKwInOut{Input}{Input}
\SetKwInOut{Output}{Output}

\Input{Text prompt $T$}
\Output{Optimized SVG canvas $C^*$}

\textbf{Layout Generation and Correction}\;
Generate intial layout $L_i$ via LLM \cite{wu2023self}\;
%Initialize $E \gets \infty$\;
%\While{$E > \epsilon$}{
%  $L \gets \text{LLM}(T)$ \tcp*{Boxes $(b_i, \tau_i)$}
%  $\hat{T} \gets \text{Reconstruct}(L)$\;
%  $E \gets \lambda_1 \text{CosSim}(T, \hat{T}) + \lambda_2 \text{Jacc}(T, \hat{T}) + \lambda_3 \text{Lev}(T, \hat{T})$\;
%}
set $i \gets 1$; $\delta_{rec}^{i-1} \gets 1$; $\delta_{rec}^i \gets 10e9+7$;\\
\Repeat{$\delta_{rec} \neq 0$ \text{and} $|\delta_{rec}^{i-1}-\delta_{rec}^i|<1e-4$}{
    reconstruct caption with \cite{yin2017obj2text};\\
    update $\delta_{rec}^i$ as defined in \cref{eq:1};\\
    set $\delta_{rec}^{i-1} \gets \delta_{rec}$;\\
    layout reconstruction via LLM with new $\delta_{rec}$;\\
    $i \gets i+1$;
    }
final layout $L_f$

\textbf{Canvas Initialization via latent guided diffusion}\;
\ForEach{ boxes $b_i \in L_f$}{
  Sample $z_0 \sim \Gamma$\;
  $A_i \gets \text{softmax}(k \cdot \text{attn}(z_0, \tau_i))$\;
  %Init strokes in $b_i$ using $A_i$\;
%\textbf{Latent-Guided Diffusion}\;
\ForEach{timestep $t$}{
  $z_t \gets \text{Denoise}(z_t; \tau_i, A_i)$\;
  $z_t \gets A_i \odot z_t + (1 - A_i) \odot z_t$\;
}
Optimize the $E_c$ via \cref{eq:4}.}
Obtained the rasterized image $\mathcal{I}_r$.\\
\textbf{SVG Optimization}\;
Keypoint-based stroke initialization in $C^*$.\\
Foreground abstraction with \cref{eq:clip}.\\
Background abstraction with \cref{eq:align}.\\

Obtained $C^*$ by optimizing $\mathcal{L}_{synth}.$

\Return{$C^*$}
\label{algo1}
\end{algorithm}

%\vspace{-4mm}
\subsection{LLM-based layout generation and correction}
\label{sec:3.1}
\begin{figure}[!htbp]
%\vspace{-2mm}
\centering
  \includegraphics[width=\linewidth]{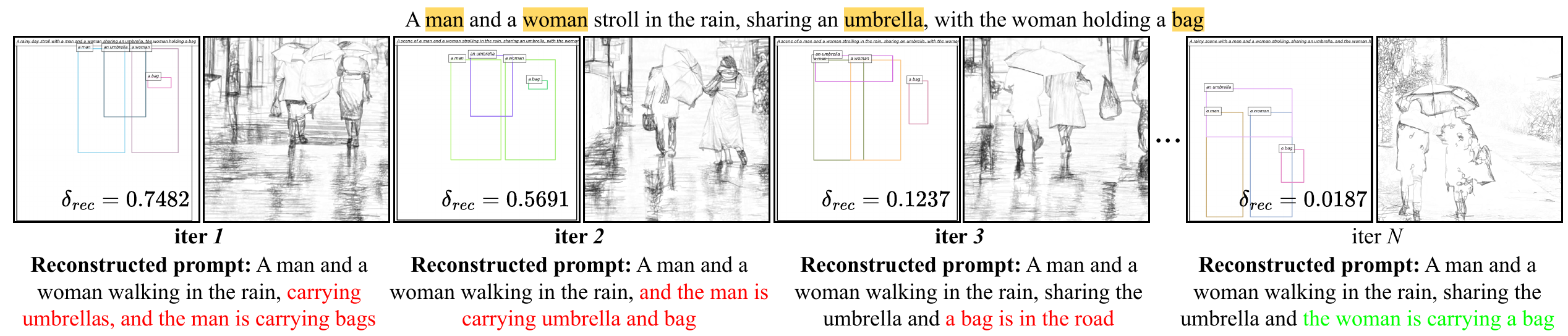}
\caption{\textbf{Iterative layout correction} via a supportive error term. %(initially $\delta_{rec} = 1$) during LLM prompting and optimize it through similarity maximization.
%(\textcolor{red}{Red}: incorrect; \textcolor{green}{Green}:closest reconstruction).
}
\label{fig:llm_correction}
%\vspace{-4mm}
\end{figure}

Leveraging layouts provides diffusion models with improved control over object enumeration, spatial arrangements, and relationships compared to text alone.
To generate a layout from the provided text prompt $T_p$, we embed $T_p$ into a template and query the LLM for completion as described in \cite{lian2023llm}. 
This process generates: (1) a concise caption and corresponding bounding boxes for foreground objects, specifying coordinates in $(x, y, w, h)$ format (see \cref{fig:llm_correction}), and (2) a background description with a negative prompt for object exclusions.
To minimize the potential of ambiguous layouts generation by the LLM, which might confuse the diffusion model, we introduce an objective function, \emph{caption reconstruction error} ($\delta_{rec}$), to refine layouts. This iterative procedure uses LLM to generate a layout until the reconstruction error $\delta_{rec}$, defined in \cref{eq:1}, meets convergence criteria ($|\delta_{rec}^{i-1} - \delta_{rec}^{i}| < \epsilon: \epsilon$ = 1e-4).
\vspace{-2mm}
\begin{equation}
\vspace{-2mm}
    \label{eq:1}
    \resizebox{0.9\columnwidth}{!}{%
    $\delta_\text{rec} = 1 - [\lambda_1 \text{sim}_\text{cos} (T_p, T_r) + \lambda_2 \text{sim}_\text{jac} (T_p, T_r)  - \lambda_3 \text{sim}_\text{edit} (T_p, T_r)]
    $}
\end{equation}
where $T_r$ denotes reconstructed text prompt via \cite{yin2017obj2text} and $\text{sim}_\text{cos}, \text{sim}_\text{jac}$ and $\text{sim}_\text{edit}$ respectively represent cosine similarity, Jaccard similarity, and Edit distance. Our objective is to maximize the $\text{sim}_\text{cos}$, $\text{sim}_\text{jac}$ and minimize $\text{sim}_\text{edit}$ to minimize $\delta_\text{rec}$ between $T_p$ and $T_r$, aligning the reconstructed caption $T_r$ with the given caption $T_p$. The regularization parameters $\lambda_1$, $\lambda_2$, and $\lambda_3$ are respectively set to 1.0, 1.0, and 0.01, determined through hyperparameter grid search. The overall iterative in-context learning framework has been obtained in Algo. \ref{algo1} (more details are in \cref{supp:02}). %For more details on the LLM prompting, please refer to the \ref{supp:02}.

%\begin{algorithm}[t]
%\resizebox{\textwidth}{!}{
%\SetAlgoNoLine
%\KwIn{User prompt $T_p$, Initial layout $L_i$} 
%\KwOut{Final layout $L_f$.}
%set $i \gets 1$; $\delta_{rec}^{i-1} \gets 1$; $\delta_{rec}^i \gets 10e9+7$;\\
%\Repeat{$\delta_{rec} \neq 0$ \text{and} $|\delta_{rec}^{i-1}-\delta_{rec}^i|>1e-4$}{
%    reconstruct caption with \cite{yin2017obj2text};\\
%    update $\delta_{rec}^i$ as defined in \cref{eq:1};\\
%    set $\delta_{rec}^{i-1} \gets \delta_{rec}$;\\
%    layout reconstruction via LLM with new $\delta_{rec}$;\\
%    $i \gets i+1$;
%    }
%\caption{Iterative layout correction}
%}
%\label{algo1}
%\end{algorithm}

In Algo. \ref{algo1}, it can be observed that we did not retrain or finetune the LLM; we asked the LLM iteratively to correct the layout via minimizing the $\delta_{rec}$. As the error term $\delta_{rec}$ is a non-convex function, this is the best way of optimizing to obtain a more accurate layout, followed by the prompt leads to a significant improvement in SVG synthesis.

\subsection{Per-box mask latent-based canvas initialization}
\label{sec:3.2}
CraftSVG initializes a canvas using $\theta$, focusing the important regions for stroke placement on foreground objects to enhance control and efficiency, avoiding unnecessary strokes and eliminating the need for complex post-processing. The canvas is initialized with B\'{e}zier curves at identified keypoints, with varying opacity controlled by the semantic-aware optimization of our method. We use per-box mask latent-based initialization, as shown in \cref{fig:sketchcraft}. For each foreground object $o_i$, we synthesize a semantic-aware attention map $A_{s}^i$ (defined in \cref{eq:2}) to guide target image generation $\mathcal{I}_r$ via diffusion U-Net, denoising from $\mathcal{Z}_T^i$ to $\mathcal{Z}_0^i$ according to the background prompt. We use a gamma noise distribution ($\Gamma$) instead of the traditional Gaussian noise, as it provides higher contrast between foreground and background objects. This noise latent is shared across all $o_i$ and the background to ensure coherent viewpoint, style, and lighting ($\mathcal{Z}_T^{o_i} = \mathcal{Z}_T^{bg_i} = \mathcal{Z}_T^i \sim \dfrac{1}{\alpha} \Gamma (\alpha, 1)$| $\alpha>1$).

\myparagraph{Canvas initialization:} %For canvas initialization 
We obtain the semantic aware attention features $A_{s}^i$ for each foreground object $o_i$ and normalize them using the softmax function with affinity division: %as shown in \cref{eq:2}.
\vspace{-3mm}
\begin{equation}
\vspace{-2mm}
    \label{eq:2}
    A_s^i = \text{softmax}(K_y^T \otimes Q_x)/d^2
\end{equation}
where $\otimes$ is the cross product between $Q_x$ (pixel query features at position $x$) and $K_y$ (keys from the background description at position $y$), with $d$ measuring the affinity distance between $Q_x$ and $K_y$ considering their latent projection dimension. 
This process maximizes marginalization over the probability distribution, which helps the model to define a clear boundary between the foreground object and the background, ensuring the object aligns with the bounding box.
We then merge these refined $A_s^i$ with $A_s^{i-1}$ (the attention map from the previous iteration) to create a unified attention map $A_s$ through iterative element-wise multiplication ($\odot$) of each foreground object's attention map: %as shown in \cref{eq:3}: 
\vspace{-2mm}
\begin{equation}
\vspace{-2mm}
\label{eq:3}
A_s^k = \begin{cases} 
A_s^1 & \text{if } k = 1, \\
A_s^{k-1} \odot A_s^k & \text{if } 2 \leq k \leq N.
\end{cases}
\end{equation}
where $N$ is the number of bounding boxes generated by LLM, and $A_s^k$ represents the cumulative attention map up to the $k$-th term. The final attention map, $A_s$, is given by $A_s^N$. This composite attention map $A_s$ forms the basis for selecting $\eta$ keypoints for stroke / B\'{e}zier curve initialization in $\mathcal{C}$. %($\eta$ is user-defined, an ablation study on its effect is shown in \cref{fig:ab_strokes} of the Appendix). 
We use magnitudes from the attention-derived distribution map as weight parameters to guide the selection process, ensuring stroke positions are influenced by salient features. This semantic-aware selection eliminates unnecessary strokes without post-processing, enabling faster convergence and more realistic drawings.

\myparagraph{Guided image synthesis:}
We transfer the semantic aware attention maps $A_s^i$ from per-box generation through the U-Net for composed generation of the corresponding regions with semantic guidance adapting the energy function $E_c (A_s, A_s^i, i, u)$, defined in \cref{eq:4}:
%
%\vspace{-6mm}
\vspace{-2mm}
\begin{equation}
\vspace{-2mm}
\label{eq:4}
\resizebox{0.9\columnwidth}{!}{$
E_c (A_s, A_s^i, i, u) = -\text{Topk}_u(A_s^i \odot \mathcal{Z}_T^{o_i}) + \beta \text{Topk}_u(A_s^i \odot \mathcal{Z}_T^{bg_i}) +  \gamma\sum_{i = 1}^k\left|\dfrac{A_s^i - A_s}{\sigma_{A_s^i}}\right|$}
\end{equation}
where $\text{Topk}_u$ computes the mean of top-k pixel values across the spatial dimension $u$. $\beta$ and $\gamma$, set to 7.5 and 3.5, respectively, balance image coherency and semantic control. $\sigma_{A_s^i}$ represents the standard deviation of pixels in the homogeneous region $A_s^i$, controlling the covariance of that region. We use \textit{Mahalanobis distance} instead of absolute or Euclidean distance to better handle the occlusion, resulting in more complete synthesis compared to \cite{lian2023llm,qu2023layoutllm}. At each denoising timestep $t \in T$ we place each masked foreground latent $\mathcal{Z}_T^{o_i}$ onto the background latent $\mathcal{Z}_T^{bg_i}$ replacing the original pixel content in the masked region by minimizing the $E_c$ to maintain foreground generation consistency in $\mathcal{I}_r$.

\subsection{Stroke abstraction with opacity modulation}
\label{sec:3.3}
We train an 11-layer MLP network ($MLP_s$) with vector embeddings ($\mathcal{S}_1$) 
of $\eta$ strokes through a frozen CLIP-ViT image encoder, returning the abstract structure of the foreground object $\mathcal{S}_{fg}$ by optimizing the L1 distance (for accurate foreground stroke positioning) between the CLIP embeddings of $A_s$ (\ie, $\mathcal{S}_1=CLIP(A_s)$) and $\mathcal{S}_{fg}$: 
\vspace{-2mm}
\begin{equation}
\vspace{-2mm}
    \label{eq:clip}
    \mathcal{L}_{MLP_s} (\mathcal{S}_1, \mathcal{S}_{fg}) = ||\mathcal{S}_1 - \mathcal{S}_{fg}||
\end{equation}
For background abstraction, another 11-layer MLP network ($MLP_d$) is trained in parallel to extract an $s$-directional vector ($s << \eta$) $\mathcal{D} = \{d_i\}_{i=1}^s$ from the CLIP embeddings of $\mathcal{I}_r$ (\ie, $\mathcal{S}_2=CLIP(I_r)$) to represent the probability of the $i$-th stroke appearing in the rendered background $S_{bg}$. For probabilistic stroke simplification, we multiply the width of each stroke $w_i$ by $d_i$, hiding strokes with very low probability due to their small width. This network is optimized through a simple cross-entropy loss $\mathcal{L}_{CE}$ (to classify whether the strokes belong to the background or not). Balancing $S_{fg}$ and $S_{bg}$ is essential for achieving recognizable sketches with varying degrees of abstraction and optimizing the overall canvas $\mathcal{C}_\text{combined}$ to get the final SVG $\mathcal{C}_\text{final}$. To achieve this, we define a perceptual alignment loss:
\vspace{-3mm}
\begin{equation}
\vspace{-2mm}
    \label{eq:align}
    \mathcal{L}_\text{align} (\mathcal{S}_2, \mathcal{S}_{fg}, \mathcal{S}_{bg}) = \left|\dfrac{\mathcal{S}_2 - \mathcal{S}_{fg}}{\mathcal{S}_2 - \mathcal{S}_{bg}}\right|^2
\end{equation}
The differential rendering $\mathcal{R}_d$ is trained with the loss in \cref{eq:align} to obtain $\mathcal{C}_\text{combined}$, which is further optimized with LPIPS loss and opacity modulation to obtain the $\mathcal{C}_\text{final}$. 

\myparagraph{Semantic aware opacity enhancement:}
CraftSVG aims to replicate the iterative approach of traditional human sketching by starting with a low opacity value for the B\'{e}zier curves. The system then increases the opacity of certain strokes/primitives based on relevance and significance provided by the semantic guidance through $\mathcal{L}_{sop}$: %as defined in \cref{eq:5}.
\vspace{-2mm}
\begin{equation}
\vspace{-2mm}
    \label{eq:5}
    \mathcal{L}_\text{sop} (\mathcal{S}_1, \mathcal{I}_r, \theta) =  \left|1 - \dfrac{\max(\mathcal{S}_1 \odot \mathcal{R}_d(\theta))}{\max(\mathcal{S}_1 \odot \mathcal{I}_r)}\right|
\end{equation}
The system updates the opacity value of the strokes in each backward pass through gradient descent to mirror the artistic method of intensifying strokes. We have taken the max values from both the final product matrices as we aim to intensify a certain amount of strokes in our region of interest.

\myparagraph{Optimization criteria:}
Our optimization criteria is to maximize the similarity between the rasterized target image $\mathcal{I}_r$ and the synthesized SVG parameters $R_d (\theta)$ through LPIPS loss \cite{zhang2018unreasonable}. To measure this similarity, we utilize a pre-trained image encoder $\mathcal{I}$ for feature encoding from both $\mathcal{I}_r$ and $R_d (\theta)$. To amplify the perceptual similarity between $\mathcal{I}_r$ and $R_d(\theta)$, we define the optimization strategy as follows:
\vspace{-2mm}
\begin{equation}
\vspace{-2mm}
    \label{eq:06}
    \mathcal{L}_\text{synth} = \text{LPIPS}(\mathcal{I}_r, R_d (\theta)) + \lambda_\text{sop} \mathcal{L}_\text{sop} + \lambda_\text{align} \mathcal{L}_\text{align}
%    \vspace{2pt}
\end{equation}
Here, $\lambda_\text{sop}$ is the weight regularization hyper-parameter set to 0.3 empirically to maintain the coherency between $\mathcal{I}_r$ and $R_d (\theta)$. On the other hand, $\lambda_\text{align}$ controls the disentanglement of abstraction between simplicity and recognizability and is set to 0.5 empirically. This $\mathcal{L}_\text{synth}$ loss function ensures that the synthesized SVGs align with the inherent semantics and perceptual details of $\mathcal{I}_r$ from the diffusion U-Net, enhancing the realism and visual appeal.

%% file: tex/04_expts.tex
\section{Experiments}
\label{sec:exp}
\begin{figure}[!htbp]
\vspace{-2mm}
\centering
\includegraphics[width=\columnwidth]{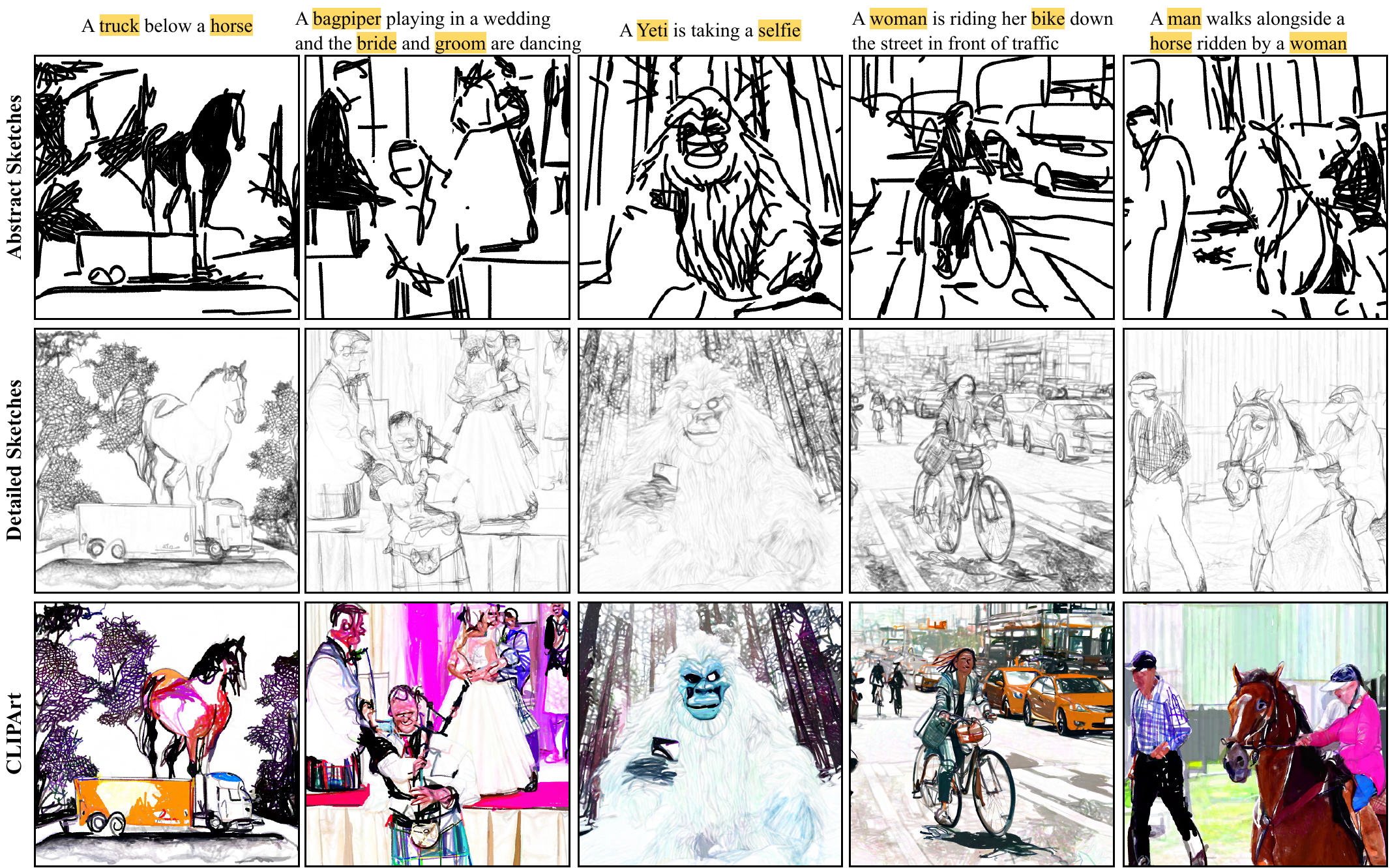}
\caption{\textbf{SVGs with CraftSVG: } 1st row depicts the abstract SVG (\#strokes = 64) obtained via the two parallel MLPs. 2nd row, the (\#strokes) = 1024 via optimizing opacity, and semantic awareness. 3rd row further optimizes the color to produce CLIPArt.}
\label{fig:viz}
\vspace{-2mm}% Give a unique label
\end{figure}

%This section illustrates the potential of CraftSVG to synthesize SVG using B\'{e}zier curves. We also compare CraftSVG with prior works and assess its components through a comprehensive ablation study to confirm their effectiveness. Some qualitative examples synthesized through CraftSVG have been depicted in \cref{fig:viz}. It has been observed that the obtained sketches are quite complete even when we abstractify them with a smaller number of strokes. Also, the abstract sketches successfully convey the information mentioned in the text prompt, which depicts the true potential of the proposed CraftSVG architecture. 
CraftSVG showcases its ability to synthesize SVGs using Bézier curves in \cref{fig:viz}. Even with fewer strokes, the abstract sketches remain complete and effectively convey the text prompts. We also compare with prior work and validate component effectiveness through ablation studies.

\begin{figure}[!ht]
\centering
%\vspace{-2mm}
\includegraphics[width=\columnwidth]{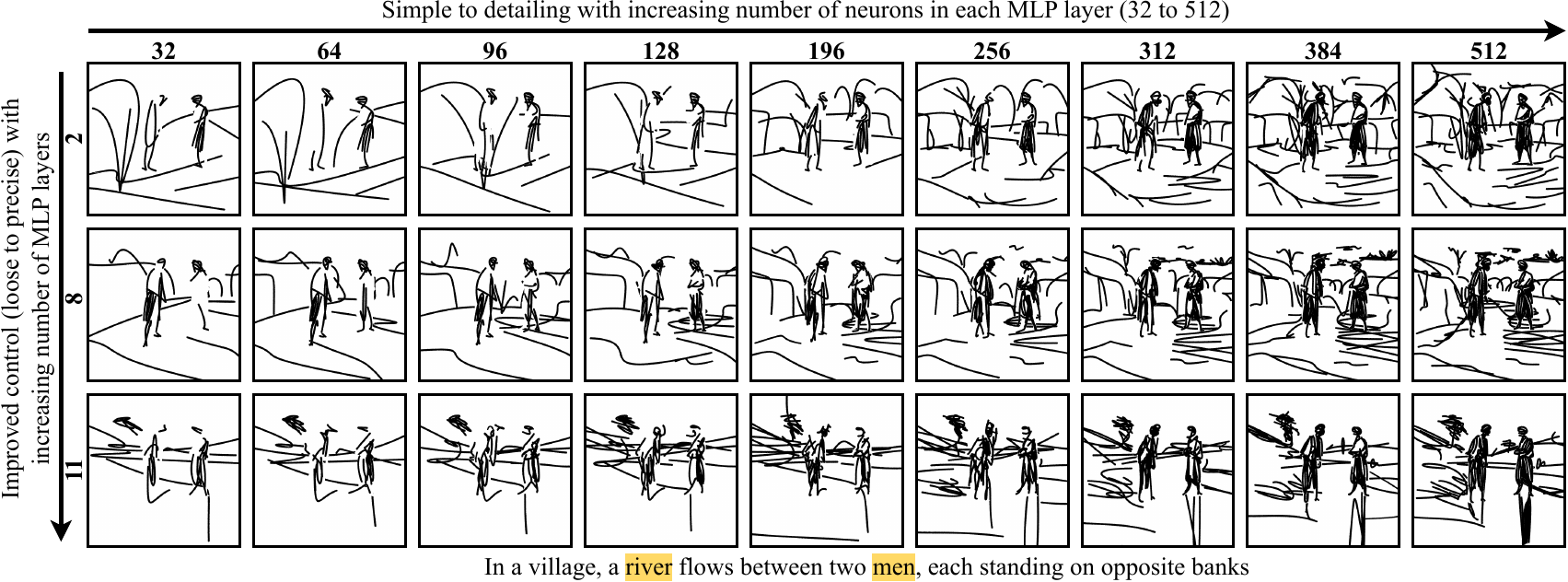}

\caption{\textbf{Strokes Abstraction:} Increasing the no. of layers and neurons enhances canvas control, detail, and aesthetics, while fewer neurons and layers maintain simplicity and recognizability (No. of strokes in MLP-based abstraction is in the range of [32, 128]).}
\label{fig:exp}       % Give a unique label
%\vspace{-4mm}
\end{figure}

\myparagraph{SVG Abstraction:}
Abstracting SVGs is crucial for effective communication, particularly during design conceptualization. This experiment demonstrates how varying the number of MLP layers and neurons in our $\text{MLP}_d$ and $\text{MLP}_s$ controls the degree of abstraction. We synthesize vector graphics with B\'{e}zier curves using iterative optimization of two parallel MLP networks with 2, 8, and 11 layers, each containing 32 to 512 neurons along with the SVG parameters, as shown in \cref{fig:exp}. We can observe that even with just a 2-layer MLP, CraftSVG can depict concepts, though the strokes are scattered and loosely correlated (\eg, the house is not well-defined). Increasing the number of neurons enhances stroke correlation but still leaves some scatter (\eg, the tree remains poorly defined). By increasing the number of layers, we achieve more precise stroke control, reducing scatter and enhancing both accuracy and aesthetics, as seen in the denser, more cohesive strokes in the last row of \cref{fig:exp}.

\begin{figure}[!htbp]
\vspace{-2mm}
\centering
  \includegraphics[width=\columnwidth]{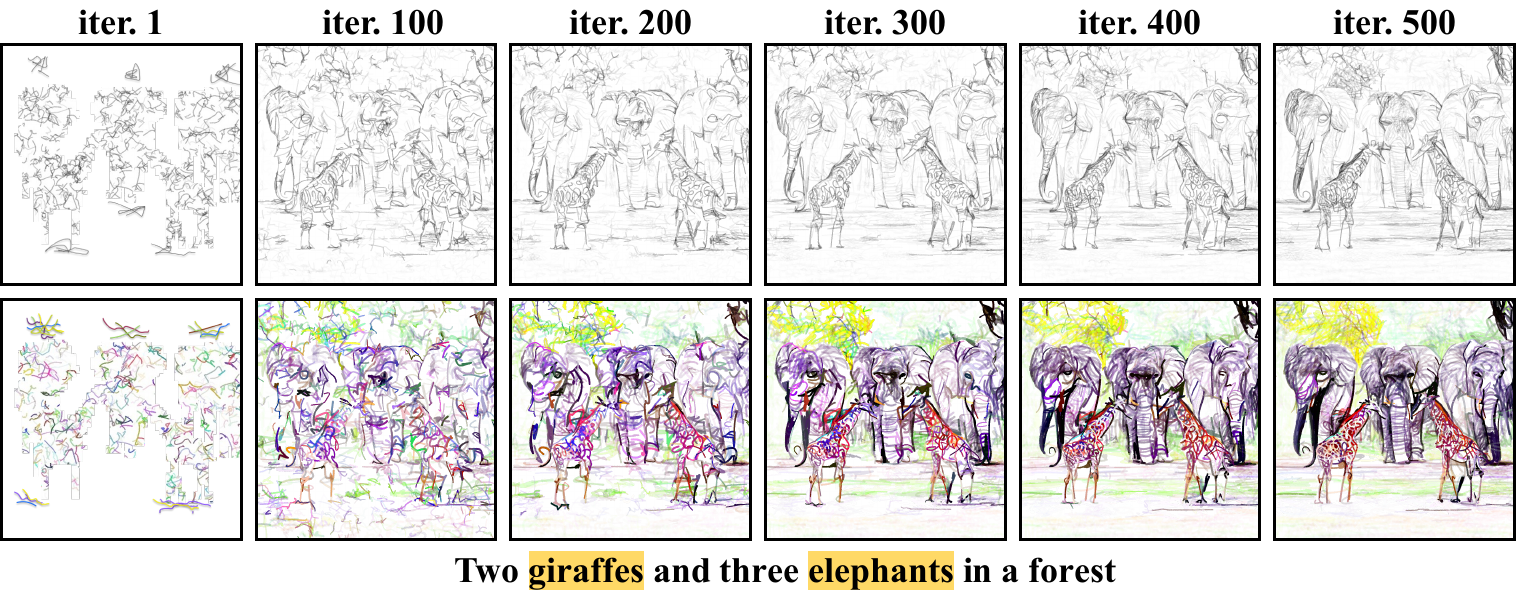}
\caption{\textbf{Evolution of Bézier curves:} A study of the affect of projective transformations in stroke deformation over the optimization iterations. (\#strokes = 1024).}
\label{fig:exp1}       % Give a unique label
\vspace{-2mm}
\end{figure}

\myparagraph{SVG synthesis via optimization:}
CraftSVG synthesizes aesthetic SVGs with B\'{e}zier curves and simple primitives via iterative optimization, aiming to explore their evolution under projective/affine transformation to obtain the final SVG as demonstrated in \cref{fig:exp1}. Here, densely initialized strokes with per-box mask latent attention (iter. 1 in \cref{fig:exp1}) are going through the iterative canvas ($\mathcal{C}$) optimization via $\mathcal{L}_\text{synth}$ in order to obtain the final SVG.
It has been observed that B\'{e}zier curves, capable of projective transformations (with 8 dof) including translation, scaling, rotation, and morphing, produce more visually appealing SVGs.
Conversely, canvas initialized with simple primitives containing circles, lines, and semi-circles (third row of \cref{fig:exp1}) results in messier, less appealing SVGs due to the limitations of affine transformations without morphing.

Morphing optimizes B\'{e}zier curves by interpolating control points for dynamic and visually appealing compositions. 
However, morphing primitives can unpredictably distort their forms, diverging significantly from their original shapes and contradicting explainability as defined in \cite{mathur2023clipdrawxprimitivebasedexplanationstext}. 
Our proposed canvas initialization strategy employs B\'{e}zier curves with single control points identified from keypoints (see \cref{fig:attention}), offering greater transformation flexibility compared to primitives, which require at least two control points, thereby limiting their adaptability (\ie, more control points imply reduced flexibility).

\subsection{Qualitative Evaluation}
\begin{figure*}[!htbp]
\centering
  \includegraphics[width=\textwidth]{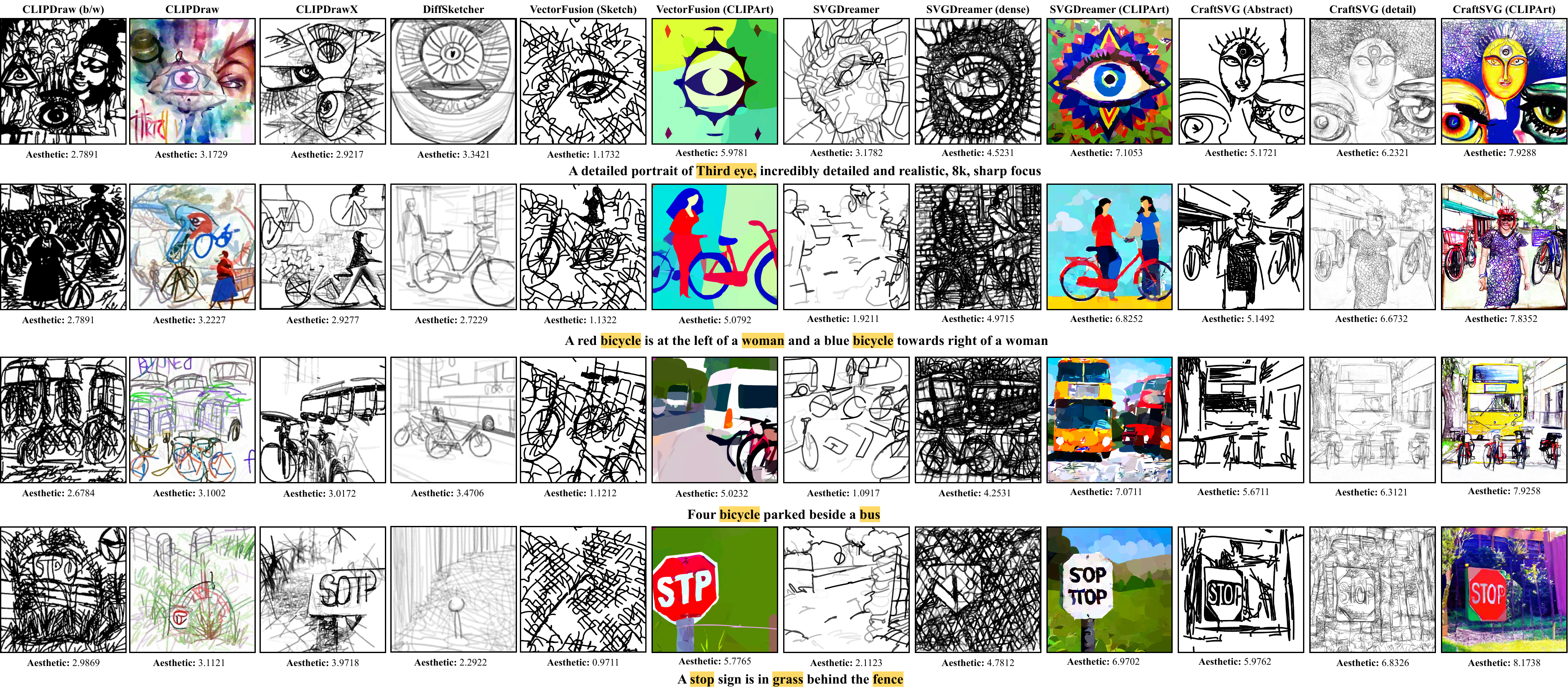}
%\Description[]{}
\caption{CraftSVG outperforms the prior works by maintaining the enumeration and spatial relationship described in the prompt. (For fair comparison, \#strokes = 1024 for sketch and CLIPArt for all the methods. For Abstract SVGDreamer and CraftSVG, the \#strokes = 128.)}
\label{fig:comp_study}       % Give a unique label
\vspace{-4mm}
\end{figure*}

We compare CraftSVG with the CLIP-based method CLIPDraw \cite{frans2022clipdraw}, CLIPDrawX \cite{mathur2023clipdrawxprimitivebasedexplanationstext} as well as diffusion-based techniques like DiffSketcher \cite{xing2023diffsketcher}, VectorFusion \cite{jain2023vectorfusion}, and SVGDreamer \cite{xing2023svgdreamer}. As shown in \cref{fig:comp_study}, CraftSVG excels in producing SVGs better aligned with text prompts, achieving better stroke alignment, abstraction, and opacity modulation, which adds shadow and depth. In contrast, diffusion-based methods suffer from color oversaturation, over-smoothing, and loss of fine details due to SDS loss. Additionally, CraftSVG leads in perceptual quality, as measured by the Aesthetics \cite{Aes2022}, surpassing existing methods.

In \cref{fig:comp_study}, CLIPDraw \cite{frans2022clipdraw} generates messy drawings with low aesthetic scores and fails to follow the enumeration and spatial relationships. %in text prompts. 
Although adding color slightly improves its aesthetics, still struggles with imaginary concepts like ``Third eye''. DiffSketcher \cite{xing2023diffsketcher} improves aesthetics but fails to handle multiple objects or specific scene text (\eg, row 4 in \cref{fig:comp_study} shows missing letters in ``stop sign''). VectorFusion \cite{jain2023vectorfusion} produces unrecognizable, messy drawings due to optimization in the latent space of a pixel-based diffusion model, leading to domain gap issues. Adding color improves VectorFusion’s aesthetics but results in overly saturated images lacking depth and shadow, and it fails to maintain object counts and spatial relations or accurately render text symbols (``STP'' instead of ``STOP''). SVGDreamer \cite{xing2023svgdreamer} suffers from similar domain gap problems, leading to incomplete canvases with fewer strokes and messy backgrounds with more strokes. Though color enhances its aesthetics, it still struggles with enumeration, spatial relationships. CraftSVG consistently delivers clean SVGs with accurate semantic details, reducing noise with per-box mask latent-based initialization and providing depth and shadowing with semantic-aware opacity optimization, requiring minimal manual intervention and parameter tuning.

\subsection{Quantitative Evaluation}
%We have conducted experiments to validate our approach using various metrics, including Cosine Similarity (CS) \cite{huang2024t2i} FID \cite{heusel2017gans}, PSNR \cite{hore2010image}, CLIP-T Score \cite{radford2021learning}, BLIPScore \cite{li2022blip}, Aesthetic score \cite{Aes2022}, HPSv2 \cite{wu2023human}, and confusion score \cite{xing2023diffsketcher}, comparing it with most representative text-to-SVG methods like CLIPDraw, CLIPDrawX \cite{mathur2023clipdrawxprimitivebasedexplanationstext} DiffSketcher \cite{xing2023diffsketcher}, VectorFusion \cite{jain2023vectorfusion}, and SVGDreamer  \cite{xing2023svgdreamer}.

This experiment has considered four styles -- sketch, CLIPArt, abstraction, and primitive -- each with 50 unique prompts and generating 10 SVGs per prompt. We have evaluated SVG diversity using diffusion U-Net generated images as ground truth, computing FID and PSNR metrics (see \cref{tab:01}). To assess the consistency between synthesized SVGs and text prompts, we used Cosine Similarity, CLIP-T, and BLIP scores. Perceptual quality has been assessed through the LAION aesthetic classifier \cite{Aes2022} and further evaluated human aesthetic appeal via HPS \cite{wu2023human} and a confusion score \cite{xing2023diffsketcher} through a human study, which are reported in \cref{fig:users}. Additionally, we measured memory usage during inference and calculated the hit ratio by generating 10 images from 10 different seeds per prompt to evaluate how consistently the SVGs align with the text prompts.

\begin{table}[!htbp]
%\vspace{-2mm}
\centering
\caption{Evaluation on conventional SVG synthesis techniques.}
\resizebox{\columnwidth}{!}{
\begin{tabular}{@{}lcccccccccc@{}}
\toprule
Method / Metric        & CS $\uparrow$   & FID $\downarrow$ & PSNR $\uparrow$ & CLIP-T $\uparrow$ & BLIP $\uparrow$ & Aes. $\uparrow$ & HPSv2 $\uparrow$ & Conf. $\uparrow$ & Mem. (GB) $\downarrow$ & Hit \% $\uparrow$ \\ \midrule
CLIPDraw (b/w)         & 0.2882          & 172.67           & 4.13            & 0.1886            & 0.2672          & 3.2803          & 0.1883           & 0.27             & \textbf{1.4}           & 6.8               \\
CLIPDraw (CLIPArt)     & 0.2911          & 171.17           & 4.66            & 0.1952            & 0.2704          & 2.9182          & 0.1892           & 0.27             & \textbf{1.4}           & 7.1               \\
CLIPDrawX              & 0.3276          & 146.12           & 5.92            & 0.2102            & 0.3031          & 3.2144          & 0.1997           & 0.32             & 8.4                    & 17.2              \\ \midrule
VectorFusion (Sketch)  & 0.4211          & 127.44           & 8.14            & 0.2719            & 0.2991          & 1.1671          & 0.2007           & 0.17             & 16.2                   & 7.7               \\
VectorFusion (CLIPArt) & 0.5117          & 97.94            & 11.27           & 0.3710            & 0.4421          & 5.5312          & 0.2617           & 0.33             & 16.2                   & 28.3              \\
DiffSketcher (Sketch)  & 0.3629          & 120.04           & 8.38            & 0.2855            & 0.3987          & 4.2468          & 0.2411           & 0.60             & 12.7                   & 27.6              \\
DiffSketcher (CLIPArt) & 0.3538          & 121.77           & 10.12           & 0.2892            & 0.3921          & 4.6481          & 0.2422           & 0.64             & 12.7                   & 27.8              \\ \midrule
SVGDreamer (Abstract)  & 0.4412          & 67.12            & 11.67           & 0.3001            & 0.4623          & 4.8432          & 0.2685           & 0.46             & 32.7                   & 17.6              \\
SVGDreamer (Sketch)    & 0.5214          & 61.67            & 11.87           & 0.3211            & 0.4712          & 5.1423          & 0.2718           & 0.37             & 32.7                   & 42.7              \\
SVGDreamer (CLIPArt)   & 0.5962          & 59.13            & 14.54           & 0.3440            & 0.4972          & 6.5432          & 0.2888           & 0.31             & 32.7                   & 48.8              \\ \midrule
CraftSVG (Abstract)    & 0.6176          & 51.42            & 15.98           & 0.4563            & 0.5223          & 5.9873          & 0.3167           & \textbf{0.66}    & 12.1                   & 62.1              \\
CraftSVG (Sketch)      & 0.6342          & 48.42            & 16.07           & 0.4667            & 0.5432          & 6.7832          & 0.3276           & \textbf{0.66}    & 12.1                   & 65.2              \\
CraftSVG (CLIPArt)     & \textbf{0.7091} & \textbf{39.87}   & \textbf{17.15}  & \textbf{0.5013}   & \textbf{0.5783} & \textbf{7.0779} & \textbf{0.3523}  & 0.61             & 12.1                   & \textbf{66.7}     \\
CraftSVG (Primitive)    & 0.6019          & 55.76            & 15.51           & 0.4031            & 0.5038          & 5.079           & 0.3032           & 0.54             & 12.1                   & 61.7              \\ \bottomrule
\end{tabular}
}
\label{tab:01}
\vspace{-4mm}
\end{table}

In \cref{tab:01}, CraftSVG surpasses both CLIP-based and diffusion-based approaches in all metrics, except for memory usage during inference, as loading a pre-trained diffusion model requires more memory than a pre-trained CLIP. Our method's superior FID and PSNR metrics indicate greater diversity in generated SVGs. Higher cosine similarity, CLIP-T, and BLIP scores show that CraftSVG's generated CLIPArts and sketches more closely match the given text prompts. Interestingly, the confusion score for CraftSVG's CLIPArts is lower than for sketches, likely because CLIPArts capture fine details and color contrast typical of artistic drawings, unlike free-hand human drawings, making it easier for humans to distinguish real from synthesized images. The most significant metric is the Hit ratio, with CraftSVG showing a 15\% improvement over SVGDreamer, demonstrating its consistency. As shown in \cref{fig:comp_study}, SVGDreamer and VectorFusion generate different CLIPArt and sketches for the same seeds due to a large domain gap between SVGs and pixel space, and fine-tuning the diffusion model with LoRA suffers from parameter initialization sensitivity and overfitting to low-rank subspace \cite{choi2024simple, kwon2023datainf}.

\subsection{Ablation study}
\myparagraph{Layout generation:}
This study aims to identify the optimal LLM for generating accurate layouts from text prompts that convey enumeration, position, relationships, and spatial arrangements of multiple objects. Following \cite{hong2018inferring}, we use caption generation as an extrinsic evaluation method, employing \cite{yin2017obj2text} to generate captions from semantic layouts for comparison with original scene captions. We assess performance using existing metrics like METEOR \cite{banerjee2005meteor}, ROUGE \cite{lin2004rouge}, CIDEr \cite{vedantam2015cider}, and SPICE \cite{anderson2016spice}, as detailed in \cref{tab:03}. The performances are based on 50 unique prompts tested 5 times each, without layout correction, to determine the best LLM for our work. We started testing the Text2Scene module \cite{tan2019text2scene}, which struggled with objects outside the MS-COCO dataset. We then explored LLMs, starting with Llama-7B \cite{touvron2023llama}, which improved the baseline by 8\%. Vicuna-13B \cite{chiang2023vicuna} further refined Llama-7B, offering a 5\% improvement, and GPT-3.5 surpassed Vicuna-13B by 3\%. Between GPT-4o and Claude-3.5 Sonnet, with a 2:3 ratio favoring Claude-3.5 Sonnet, we chose the latter for our experiments.

\begin{table}[!htbp]
\vspace{-2mm}
\centering
\caption{Layout generation with LLMs.}
\resizebox{\columnwidth}{!}{
\begin{tabular}{l|c|c|c|c|c}
\hline
Methods      & METEOR $\uparrow$ & ROUGE $\uparrow$ & CIDEr $\uparrow$ & SPICE $\uparrow$ & Average $\uparrow$ \\ \hline
Text2Scene   &   0.189     &    0.446   &   0.601   &    0.123   &   0.339      \\
Llama-7B    &  0.232      &   0.573    &    0.648   &    0.214   &    0.416     \\
Vicuna-13B &    0.274    &  0.656     &   0.721    &    0.222   &     0.468    \\
% GPT-3.5 \footnote[6]{https://huggingface.co/Xenova/gpt-3.5-turbo}      &     \textbf{0.271}   &    \textbf{0.711}   &    \textbf{0.753}   &   \textbf{0.254}    &    \textbf{0.497}     \\ \hline
GPT-3.5    &     0.271   &    0.711   &    0.753   &   0.254    &    0.497     \\
GPT-4o    &     0.289   &    \textbf{0.737 }  &    0.797   &   \textbf{0.382}    &    0.511     \\
Claude-2    &     0.291   &    0.717   &    0.782   &   0.291    &    0.538     \\
Claude-3.5 Sonnet    &     \textbf{0.302}   &    0.728   &    \textbf{0.801}   &   0.322    &    \textbf{0.544}    \\ \hline
\end{tabular}
}
\label{tab:03}
\vspace{-2mm}
\end{table}

\begin{figure}[!ht]
%\vspace{-2mm}
\centering
\includegraphics[width=\columnwidth]{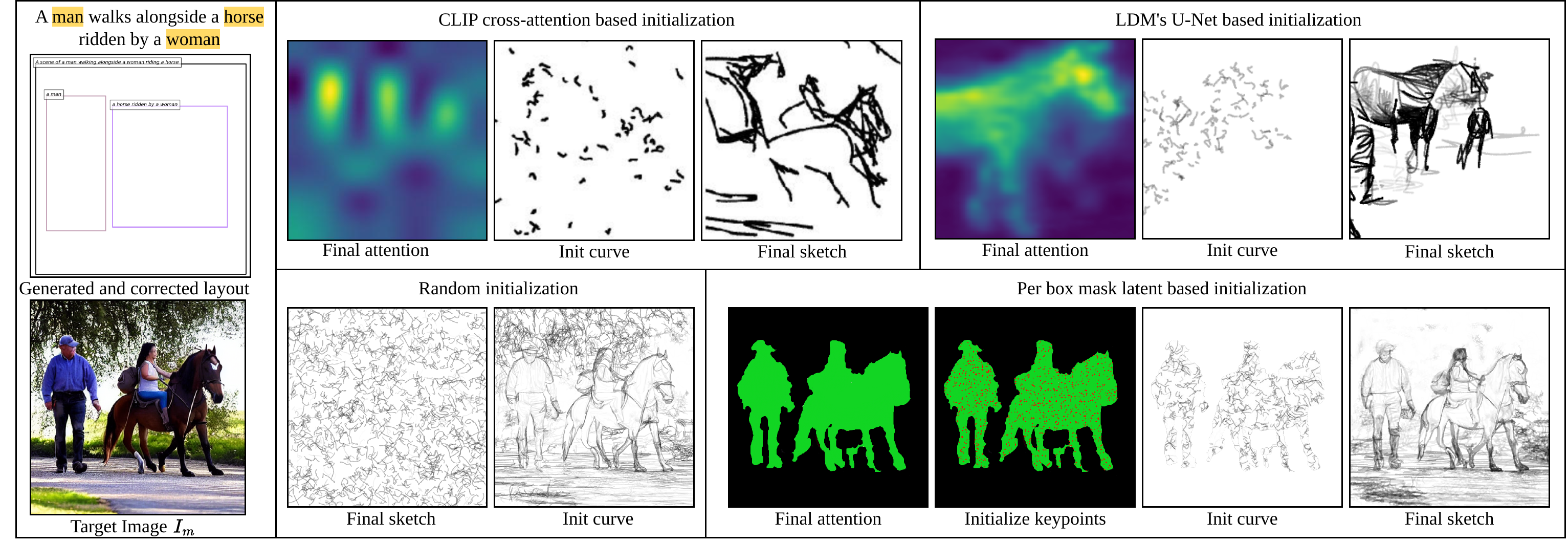}
%\Description[]{}
\caption{\textbf{Canvas Initialization:} LDM and CLIP attention fail to capture object-level detail, while random init adds clutter. Per-box mask latent attention ensures complete, detailed canvas coverage.}
\label{fig:attention}       % Give a unique label
\vspace{-4mm}
\end{figure}

\myparagraph{Canvas initialization:}
The non-convex nature of CraftSVG's optimization is sensitive to the initialization of B\'{e}zier curves or primitives, which made us examine four different initialization types: (1) CLIP-based \cite{vinker2022clipasso}, (2) diffusion U-Net-based \cite{xing2023diffsketcher}, (3) random initialization, and (4) per box mask latent-based attention (see \cref{fig:attention}). CLIP-based initialization offers global attention but lacks object precision, while diffusion U-Net-based focuses on singular objects, enhancing details but overlooking others, yielding higher aesthetic scores. Random initialization can fill the canvas but risks messiness with unnecessary strokes, underscoring the importance of semantic guidance. This led to the development of per-box mask latent-based initialization, capturing minute details, accurate enumeration and spatial relationships among multiple objects.
%However, the keypoints inside the foreground region have been initialized randomly.

\begin{figure}[!ht]
%\vspace{-2mm}
\centering
  \includegraphics[width=\columnwidth]{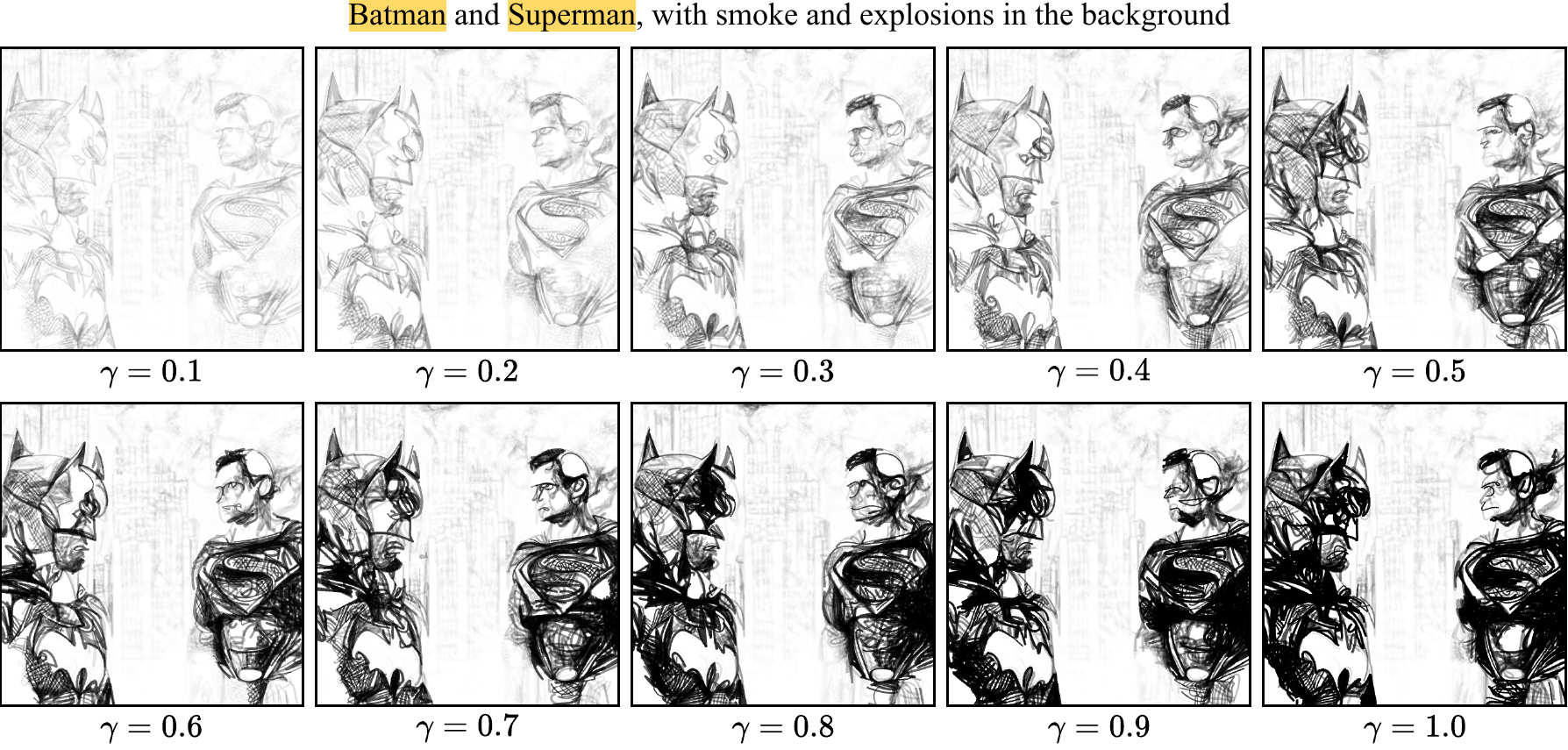}
%\Description[]{}
\caption{\textbf{Increasing opacity:} initializing strokes with higher $\gamma$ values enhances foreground-background contrast ($n$: 1024, $w$: 4.0).} %\anjan{to have space, figures with $\gamma=0.1, 0.3, 0.5, 0.8, 1.0$ could be shown}}
\label{fig:ab_opconc}       % Give a unique label
\vspace{-2mm}
\end{figure}

\myparagraph{Opacity ($\gamma$):}
The opacity parameter $\gamma$ has a significant influence on the final SVG outcome. Starting with transparent strokes (low $\gamma$ value), we gradually increased the opacity of only the essential strokes needed to match the semantics of the text prompt. As shown in \cref{fig:ab_opconc}, the final SVG initialized with lower $\gamma$ has very low contrast between foreground and background, which doesn't maintain the common human drawing style. Increasing $\gamma$ enhances this contrast, better emulating human drawing techniques.

\begin{figure}[!ht]
\vspace{-2mm}
\centering
  \includegraphics[width=\columnwidth]{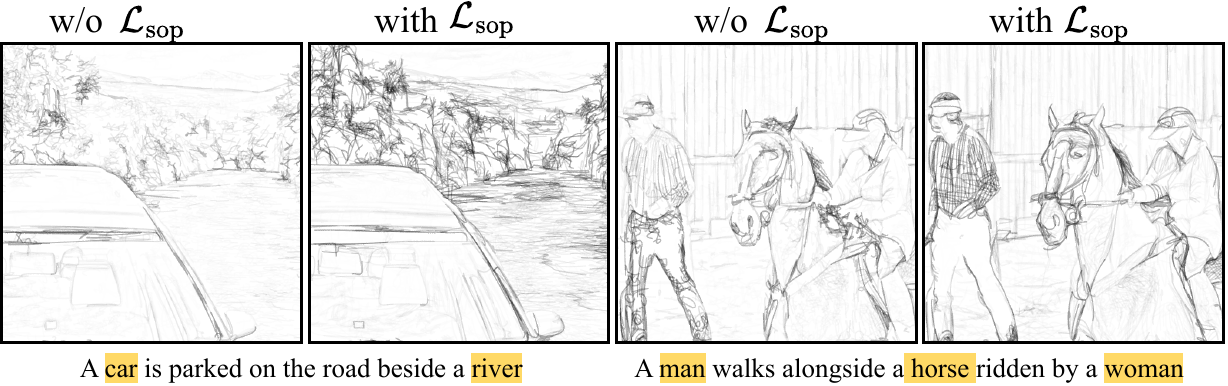}
\caption{\textbf{Semantic aware opacity} mimics human drawing by preserving symmetry, style, shading effects ($n$: 1024, $w$: 4.0, $\gamma$: 0.3).}
\label{fig:ab_op}
\vspace{-2mm}
\end{figure}

\myparagraph{Semantic aware opacity optimization:}
Opacity not only controls contrast, but also varies drawing styles, including symmetry and shading, which are crucial for emulating human-like sketches. This study highlights opacity's role by comparing SVGs with and without opacity control. If we only using LPIPS loss for similarity maximization, in the scene with "car" (see \cref{fig:ab_op}), background details like grass and reflective water strokes are absent, and in the scene with "man", "woman", and "horse", issues like asymmetrical horse ears and inaccurately drawn cap of the woman arise, diverging from human drawing style. %These examples underscore semantic aware opacity optimization's ability to enhance realism by incorporating missing features.
These examples show how semantic-aware opacity optimization adds missing details, making sketches more realistic.

\begin{figure}[!ht]
\vspace{-2mm}
\centering
  \includegraphics[width=\linewidth]{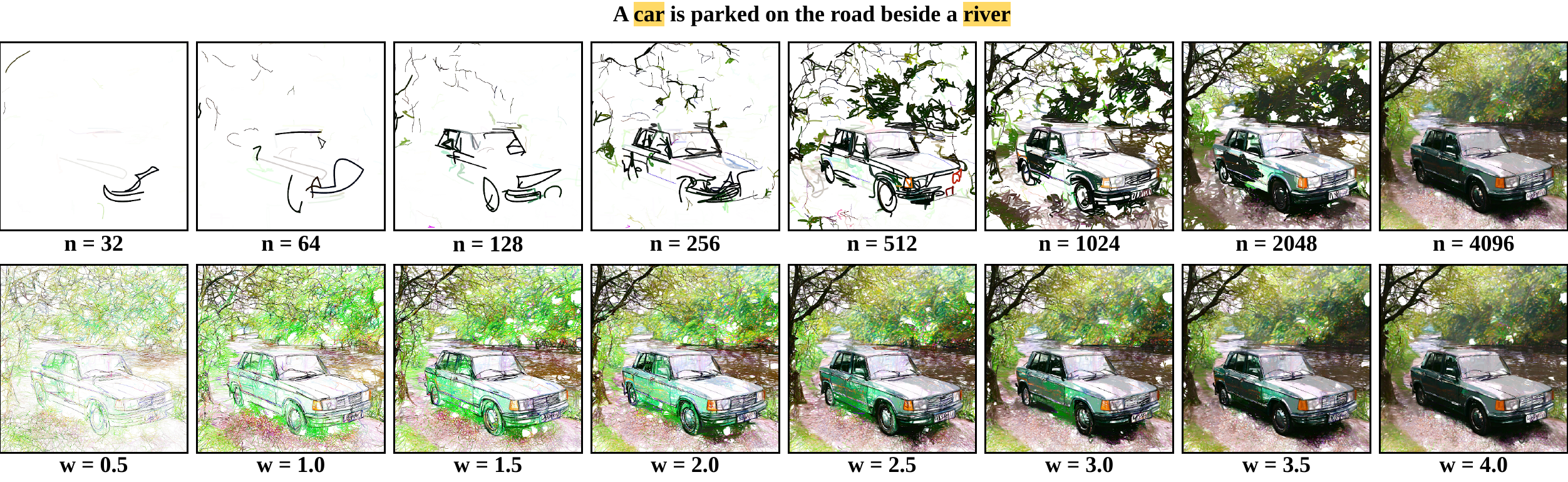}
%\Description[]{}
\caption{\textbf{No. of strokes (1st row) and its width (2nd row):} With an increasing no. of parameters, CLIPArt gets a realistic look.}
\label{fig:ab_strokes}       % Give a unique label
\vspace{-2mm}
\end{figure}

\myparagraph{Optimizing Stroke Counts and Widths:}
SVGs are all about the abstract representation of a long textual description, making it important to determine the optimal number of strokes and their width for accurately conveying a prompt.
This ablation sheds light on the optimal number of strokes and the appropriate stroke width required for drawing a certain prompt.
As shown in the top row of \cref{fig:ab_strokes}, by varying stroke numbers during initialization, our experiment found that more strokes enhance realism, but dropping below a specific threshold fails to capture the prompt's essence.
This threshold matches the canvas size, \ie, a 512 $\times$ 512 canvas requires at least 512 strokes to synthesize a realistic SVG.
Additionally, adjusting stroke width allows for various CLIPArt styles, from ``crayon effects'' with narrow widths (w = 0.5) to ``oil painting'' at wider widths (3.5-4.0), as shown in the bottom row of \cref{fig:ab_strokes}.

\subsection{Comparison with image abstraction models}
\label{supp:10}
\begin{figure}[!ht]
\vspace{-2mm}
\centering
  \includegraphics[width=\columnwidth]{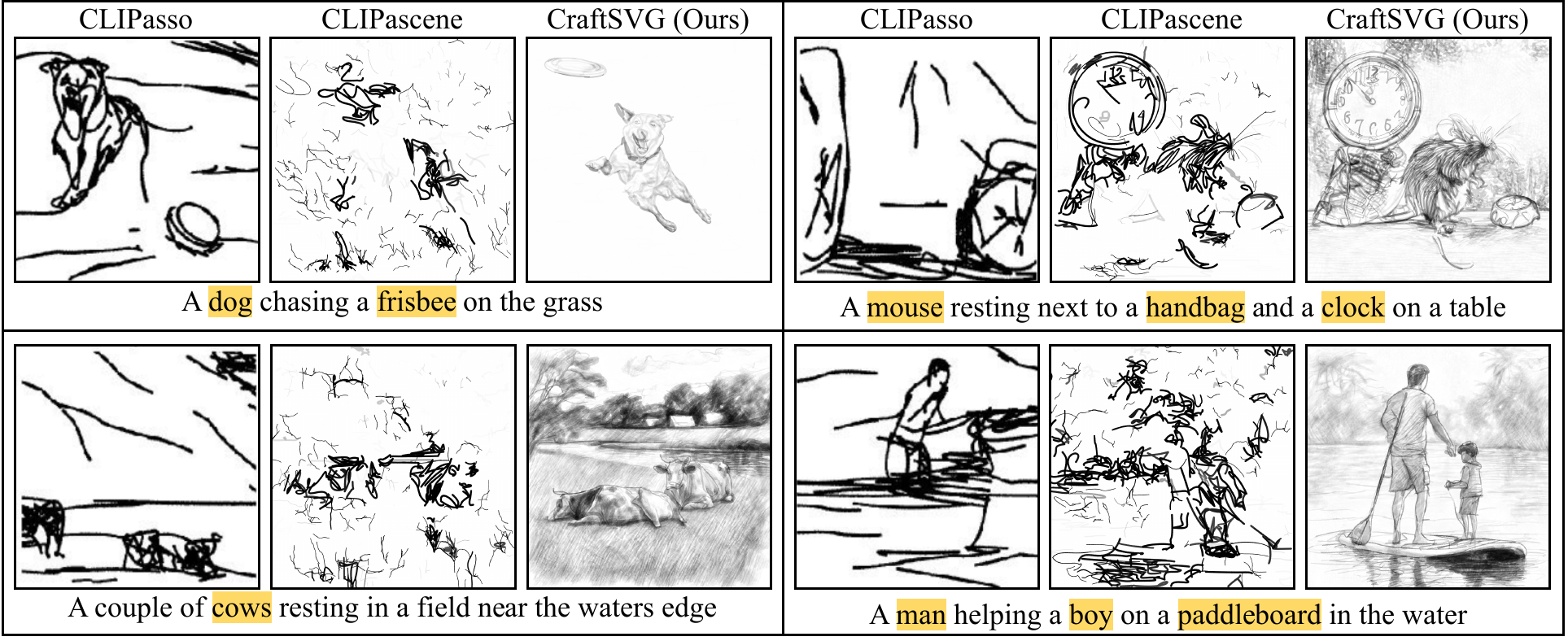}
%\vspace{-6mm}
%\Description[]{}
\caption{\textbf{Comparison on SVG Abstraction:} We have used the same rasterized images generated by the proposed diffusion method as a starting point. (\#strokes = 1024).}
\label{fig:comp_abs}       % Give a unique label
\vspace{-2mm}
\end{figure}

In CraftSVG, the canvas completion is always guided by the targeted rasterized image $\mathcal{I}_r$, so it is worth comparing CraftSVG with CLIP-based image abstraction models, namely CLIPasso \cite{vinker2022clipasso} and CLIPascene \cite{vinker2023clipascene}. As they generate vector graphics from images, we generate the target image for CLIPasso and CLIPascene with LLM-grounded diffusion, as it is the closest literature that exists concerning our work, enabling us to perform a fair comparison. The qualitative and quantitative performance of this experiment has been reported in \cref{fig:comp_abs} and \cref{supp_tab:02}, respectively.

In \cref{fig:comp_abs}, although CLIPasso can successfully generate the SVG following the text prompt ``A dog chasing a frisbee on the grass'',  it is unable to synthesize the correct SVG for the rest of the prompt. Similarly, CLIPascene successfully generated the mouse and the clock for the prompt ``A mouse resting next to a handbag and a clock on a table'', but it was unable to optimize the strokes for the rest of the prompts.

\begin{table}[!htbp]
\vspace{-2mm}
\centering
\caption{Comparison with image abstraction techniques.}
%\vspace{-2mm}
\resizebox{\columnwidth}{!}{
\begin{tabular}{@{}lccccccc@{}}
\toprule
Method / Metric &
  \begin{tabular}[c]{@{}c@{}}CS $\uparrow$ \cite{huang2024t2i}\end{tabular} &
  \begin{tabular}[c]{@{}c@{}}FID $\downarrow$ \cite{heusel2017gans}\end{tabular} &
  \begin{tabular}[c]{@{}c@{}}PSNR $\uparrow$ \cite{hore2010image}\end{tabular} &
  \begin{tabular}[c]{@{}c@{}}CLIP-T $\uparrow$ \cite{radford2021learning}\end{tabular} &
  \begin{tabular}[c]{@{}c@{}}BLIP $\uparrow$ \cite{li2022blip}\end{tabular} &
  \begin{tabular}[c]{@{}c@{}}Aes. $\uparrow$ \cite{Aes2022}\end{tabular} &
  \begin{tabular}[c]{@{}c@{}}HPSv2 $\uparrow$ \cite{wu2023human}\end{tabular} \\ \midrule
\multicolumn{1}{l}{\begin{tabular}[c]{@{}l@{}}CLIPasso \cite{vinker2022clipasso}\end{tabular}} &
  \multicolumn{1}{c}{0.3517} &
  \multicolumn{1}{c}{98.12} &
  \multicolumn{1}{c}{9.34} &
  \multicolumn{1}{c}{0.2122} &
  \multicolumn{1}{c}{0.3106} &
  \multicolumn{1}{c}{3.9523} &
  0.1991 \\ \midrule
\multicolumn{1}{l}{\begin{tabular}[c]{@{}l@{}}CLIPascene \cite{vinker2023clipascene}\end{tabular}} &
  \multicolumn{1}{c}{0.3216} &
  \multicolumn{1}{c}{101.08} &
  \multicolumn{1}{c}{8.82} &
  \multicolumn{1}{c}{0.2003} &
  \multicolumn{1}{c}{0.2937} &
  \multicolumn{1}{c}{3.0143} &
  0.1712 \\ \midrule
CraftSVG (Sketch) &
  \textbf{0.6342} &
  \textbf{48.42} &
  \textbf{16.07} &
  \textbf{0.4563} &
  \textbf{0.5223} &
  \textbf{6.7832} &
  \textbf{0.3167} \\ \bottomrule
\end{tabular}
}
\label{supp_tab:02}
\vspace{-4mm}
\end{table}

In contrast, from \cref{supp_tab:02} it can be concluded that CLIPascene performs worse than CLIPasso for the complex prompts, which doesn't support their original claim. This result occurs due to the usage of MLP, whose no. of neurons increases exponentially with increasing complexity, which is difficult to optimize further. However, this head-to-head comparison reveals the superior performance of CraftSVG in terms of textual alignment and visual quality.

\subsection{Failure cases: human face generation}
\label{supp:11}
\begin{figure}[!ht]
\vspace{-2mm}
\centering
 \includegraphics[width=\columnwidth]{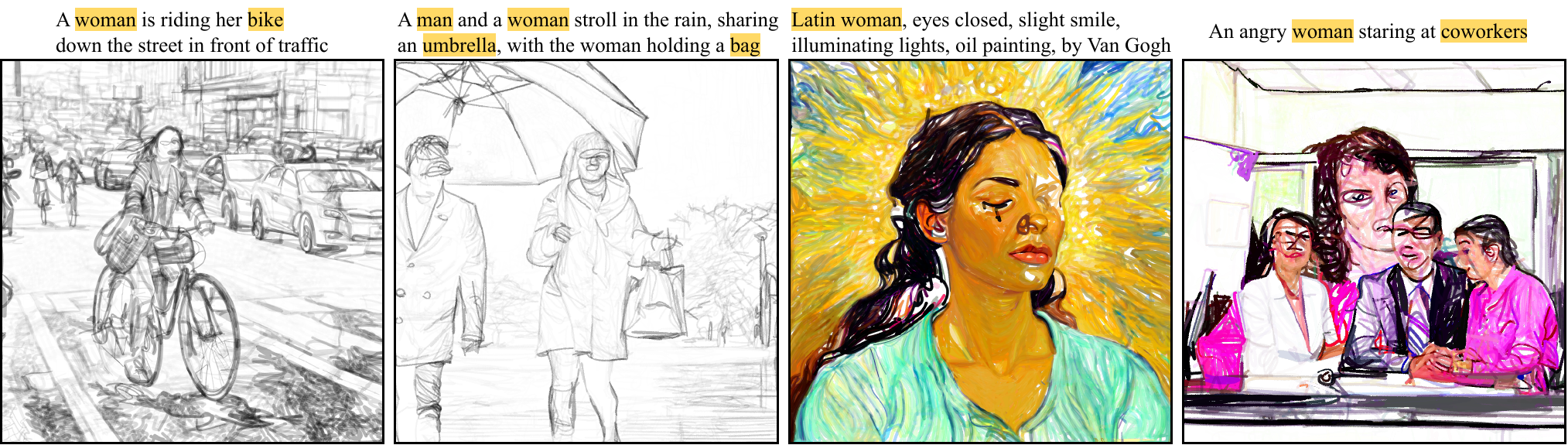}
%\Description[]{}
\caption{\textbf{Failure cases:} CraftSVG is unable to synthesize detailed faces while maintaining enumeration and spatial relationship.}
\label{fig:fail}% Give a unique label
\vspace{-4mm}
\end{figure}

CraftSVG, while adept at creating attractive SVGs with enumeration and accurate spatial relationships among multiple objects, %as directed in a text prompt, 
encounters limitations in generating precise human faces, which may be attributed to certain factors inherent to its design. First, the complexity of human faces, characterized by subtle variations in features and expressions, demands a high degree of detail that CraftSVG does not consider. With \cref{eq:06}, we optimize the canvas using criteria focusing on maximizing perceptual and semantic relevance with the reconstructed image $\mathcal{I}_r$ and does not impose any fine-grained rule to maximize precision, which is the reason our CraftSVG fails to generate human faces (see \cref{fig:fail}). Synthesizing realistic human faces would require considering fine-grained criteria focused on specific regions, which might be an interesting research direction.

%% file: tex/05_concl.tex
\section{Conclusion}
\label{sec:concl}
We proposed CraftSVG, a novel optimization-based architecture for SVG synthesis that accurately captures enumeration, relationships, spatial arrangements, and imaginative concepts. CraftSVG synthesizes free-hand vector graphics via projective transformations of B\'{e}zier curves, leveraging novel techniques such as layout correction, per box mask latent-based and semantic-aware canvas initialization, MLP abstraction, and perceptual similarity loss. These combined techniques enhance model efficiency and output high-quality SVGs, closely replicating human drawing styles. Extensive experiments and ablation studies validate CraftSVG's superiority over existing methods, showcasing its ability to produce aesthetically pleasing and semantically rich SVGs while preserving spatial relationships. However, it is unable to draw human faces properly (see \cref{fig:fail}), which we will address in our future work.

%\section{Acknowledgments}
%This work has been partially supported by the Spanish projects RTI2018-095645-B-C21, and FCT-19-15244, and the Catalan projects 2017-SGR-1783, the CERCA Program / Generalitat de Catalunya and PhD Scholarship from AGAUR (2021FIB-10010).

\section*{Acknowledgements}
This piece of research was carried out with the support of the following grant projects:
SGR Grant 2021 SGR 01559 from the Catalan Government, GRAIL PID2021-126808OB-I00, and SUKIDI PID2024-157778OB-I00 grants from the Spanish Ministry of Science and Innovation, and with the support of Cátedra UAB-Cruïlla grant TSI-100929-2023-2 from the Ministry of Economic Affairs and Digital Transformation of the Spanish Government. 

%% file: tex/06_supply.tex
\setcounter{page}{1}
\maketitlesupplementary
\section{Implementation details}
\label{supp:01}
We implemented our model using PyTorch \cite{paszke2019pytorch} and leveraged the differentiable rasterizer framework, diffvg\footnote[1]{https://github.com/BachiLi/diffvg}. We utilize the diffusion U-Net of the GLIGEN \cite{li2023gligen} model without further training or adaptation. This selected GLIGEN model used stable diffusion v1.4 from diffusers\footnote[2]{https://github.com/huggingface/diffusers} as a baseline with additional layout guidance. We set this guidance scale to 7.5, as mentioned in \cite{li2023gligen}. For the energy minimization in image synthesis, we repeat the optimization step 5 times for every object denoising step until the repetition is reduced to 1. Also, the $k$ in $\text{Topk}_u$ in \cref{eq:4} is set to 203\% of the area of the mask for each mask. The background part (second term) of \cref{eq:4} is
%weighted by $\beta$ and is set to 7.5. In addition, we add a semantic control term to maintain the image coherency, weighted by $\gamma$ is set to 3.5. Then, we decouple the scheduler for energy minimization with latent composition, and it has been conducted during the first 20 iterations of denoising to accelerate generation. 

For MLP-based abstraction, we set the number of strokes $n<=64$ in the first phase of object sketching via $MLP_s$ and train it for 500 iterations. Further, to perform the background abstraction for generating a simplified sketch, we perform 100 iterative steps of $MLP_d$. Then along with $\mathcal{L}_{align}$ we define a separate step size for sampling the features of $S_{combined}$ For simplifying the $S_{combined}$, we set this step size to be {0.3, 0.4, 0.6, 0.7} for layers {2, 7, 8, 11}, respectively for both $MLP_s$ and $MLP_d$. Each simplification step is obtained through optimizing the $\mathcal{L}_{synth}$ and backpropagating the gradient descent to both MLP Networks via differential rendering $\mathcal{R}_d$. Here, we employ the Adam optimizer with a constant learning rate of 1e-5. (NOTE: Both the MLP networks have 11 layers, and the first 5 layers consist of 512 neurons. We gradually decrease the width of the network such that the 6th and 7th layers have only 256 neurons. Similarly, the 8th, 9th, and 10th layers have only 128 neurons, and the final layers have only 64 neurons.

Now, in order to optimize the SVG parameters $\theta$, we extract the vector features from the rasterized image $I_r$ and differential renderer $\mathcal{R}_d (\theta)$ via an image encoder $\mathcal{I}$. We use the CLIP-VIT B-32 \cite{wang2023exploring}, which provides global self-attention and cross-attention along with positional embeddings. We set the iterations to 500 to optimize $\mathcal{L}_\text{synth}$ to obtain the final SVG. Now, in terms of initialization, we follow the same initialization strategy as of \cite{mathur2023clipdrawxprimitivebasedexplanationstext} in order to initialize the primitive shapes with semantic control (\ie, instead of dividing the canvas into patches, we only initialize the shapes in the foreground region obtained by per-box mask latent-based attention mechanism). For the Bézier curve, the initialization is even simpler. We just decide the number of keypoints in the foreground region and initialize a single curve from each of the keypoints with a varying number of control points (we keep it random to provide more flexibility). Also, it has to be noted that, we didn't perform any augmentations mentioned in \cite{frans2022clipdraw,mathur2023clipdrawxprimitivebasedexplanationstext} as the CLIP ViT B-32 \cite{wang2023exploring} model takes care of perspective distortions which also helps us to provide faster convergence than \cite{frans2022clipdraw,mathur2023clipdrawxprimitivebasedexplanationstext,xing2023diffsketcher,jain2023vectorfusion,xing2023svgdreamer}. All the experiments has been performed in a single NVIDIA A6000 Ada generation 48GB GPU. (Note: as CraftSVG is a training-free approach, we haven't used any large-scale dataset like StarVector \cite{rodriguez2025starvector}. Also, as our main focus is multi-object SVG generation, we exclude the training-based SVG icon generation approaches \cite{wu2025chat2svg,yang2025omnisvg})

\section{Layout generation and correction}
\label{supp:02}
In every NLP task, the main challenge is to decide the prompt/ instructions we should provide the language model to solve the task. For layout generation, we divide this prompt into five sections namely, (1) Task Specification, (2) Supporting details, (3) In-context learning, (4) Layout generation, and (5) Layout correction. The details of each section are specified as follows:

\myparagraph{Task specification:}
This section describes the prompts for explaining the layout generation task to the LLM. To do that, we use the following prompt. This part is quite similar to the instructions mentioned in \cite{lian2023llm}.

\begin{figure}[!ht]
\vspace{-4mm}
  \centering
  \includegraphics[width=\columnwidth]{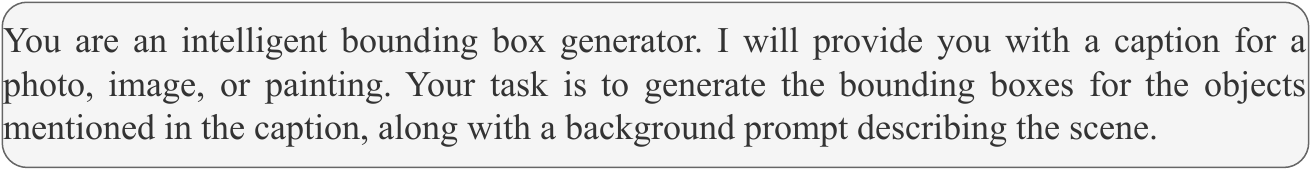}
  %%\Description[]{}
  \caption*{}
  \label{fig:tr2}
\vspace{-8mm}
\end{figure}

\myparagraph{Instruction details:}
This section mentions the details about canvas size, desired layout format, the errors we have used to optimize the layout, and so on.

\begin{figure}[!ht]
\vspace{-4mm}
  \centering
  \includegraphics[width=\columnwidth]{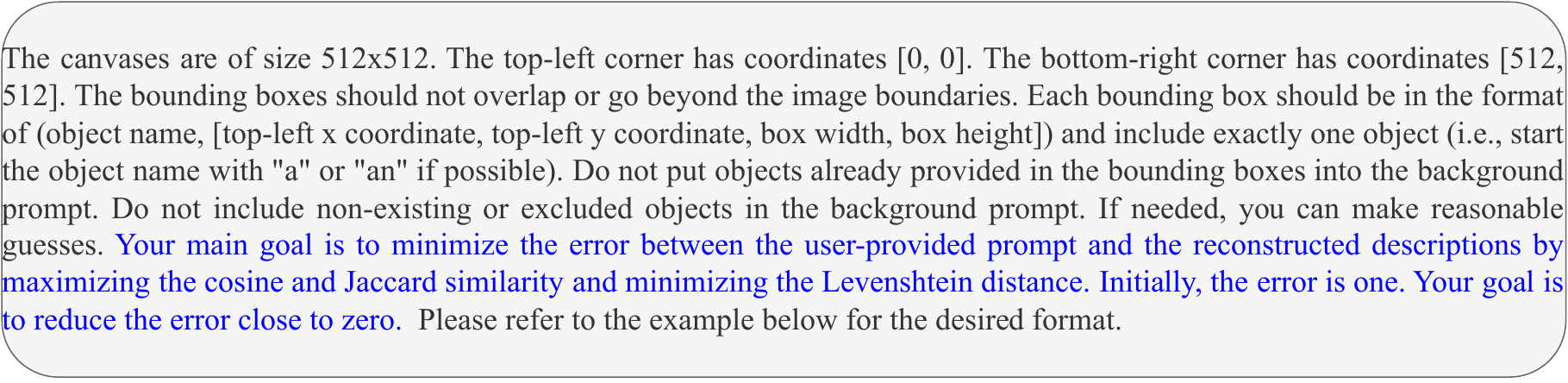}
  %\Description[]{}
  \caption*{}
  \label{fig:tr3}
\vspace{-8mm}
\end{figure}

\noindent We have used some additional prompts over \cite{lian2023llm} (highlighted in \textcolor{blue}{blue} color), which help us to enable the layout correction at a further stage.

\myparagraph{In-context learning:}
Here, we provide some basic templates, which LLM should follow during layout generation. Those examples are acquired from the MS-COCO dataset \cite{lin2014microsoft} with human specification (\ie in MS-COCO they only specified object name, here we also specified some characteristics of the object) to clarify the layout representation and obtained preferences to disperse that ambiguity. Some example templates are provided as follows:

\begin{figure}[!ht]
  \centering
  \includegraphics[width=\columnwidth]{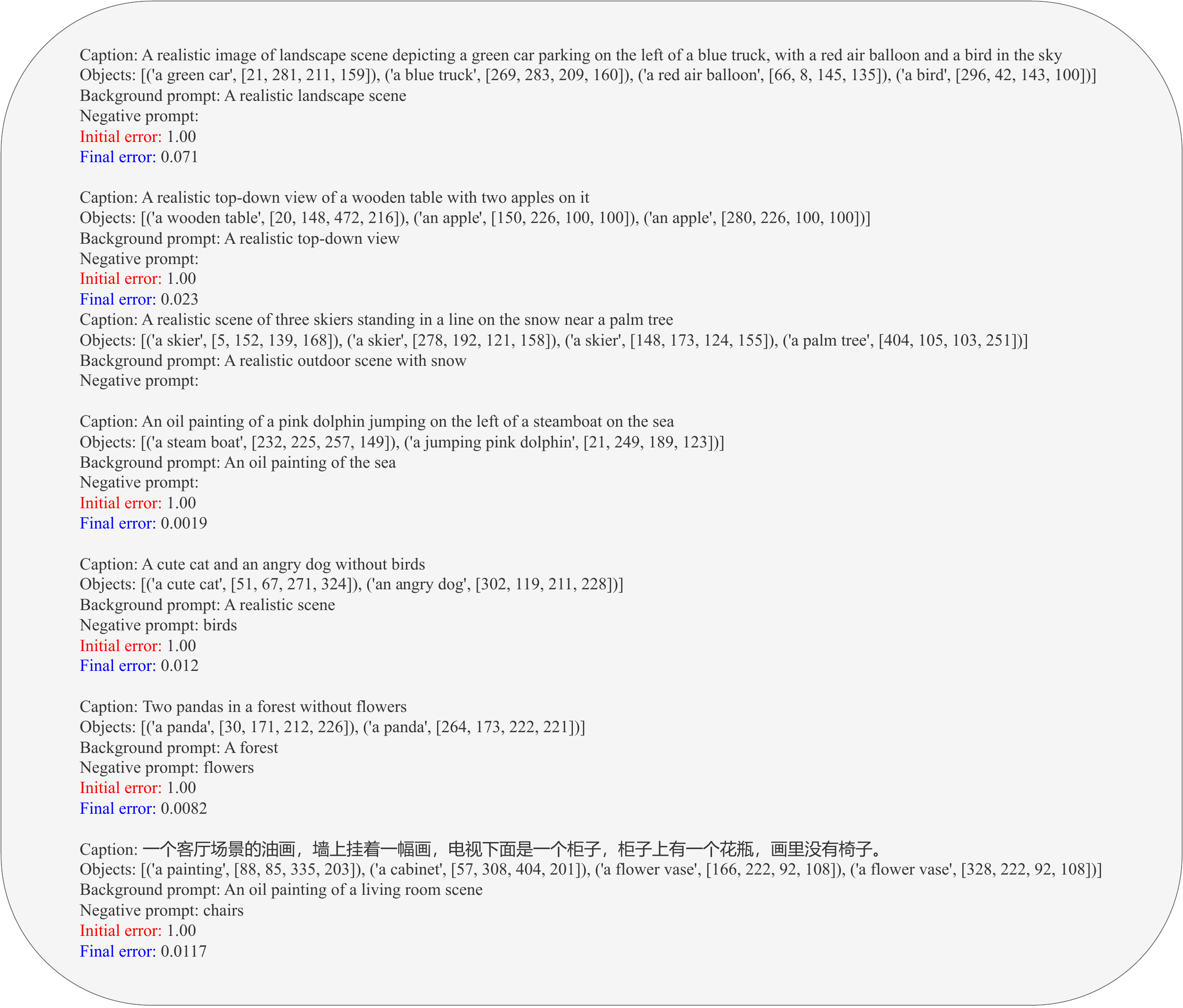}
  %\Description[]{}
  \caption*{}
  \label{fig:tr4}
\vspace{-4mm}
\end{figure}

On the top of \cite{lian2023llm}, we have mentioned our desired final error score. Based on this template, LLM generates its initial layout based on the given captions. It has to be noted that during this in-context learning, we ensured that each object
instance is represented by a single bounding box, and no foreground objects are specified in the boxes to the background caption to ensure all foreground objects contribute to the per-box mask latent-based canvas initialization.

\myparagraph{Layout generation:} This step provides interactions between users and the LLM. Here, users provide a caption, and the LLM generates its corresponding bounding boxes. An example has been provided as follows:
\begin{figure}[!ht]
\vspace{-2mm}
  \centering
  \includegraphics[width=\columnwidth]{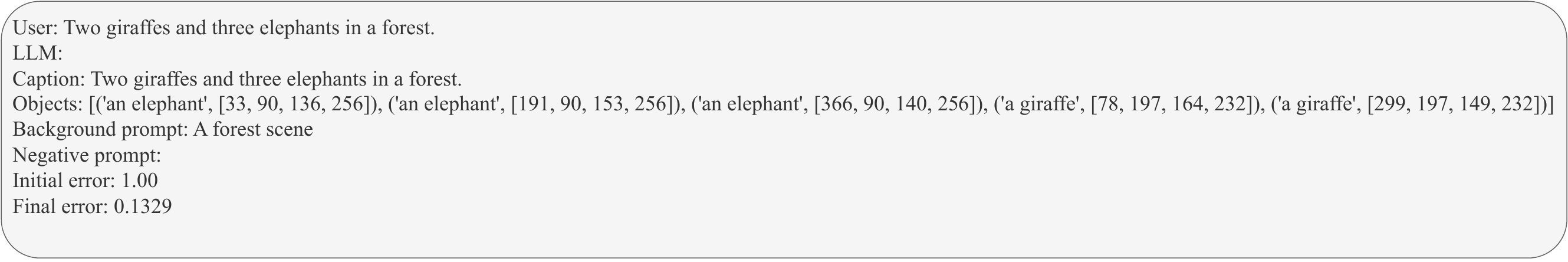}
  %\Description[]{}
  \caption*{}
  \label{fig:tr5}
\vspace{-8mm}
\end{figure}

\myparagraph{Layout correction:}
In the last step, we have performed iterative layout correction through the caption reconstruction error proposed in \cref{eq:1}. The detailed algorithm has been depicted in \cref{algo1}.

\section{Transformations affect shape deformation}
\label{supp:03}
\begin{figure}[!ht]
\centering
  \includegraphics[width=\columnwidth]{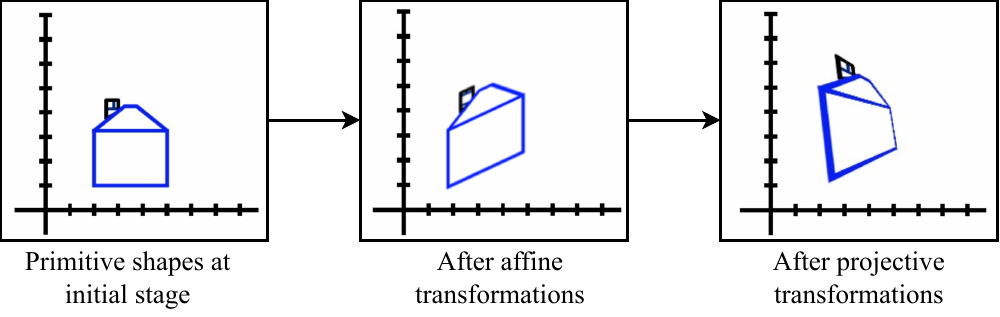}
%\Description[]{}
\caption{\textbf{Affine and projective transformations on primitive shapes:} All the primitive shapes get deformed in projective transformations via morphing and gradually lose their recognizability.}
\label{fig:ab_trans}       % Give a unique label
%\vspace{-4mm}
\end{figure}

As we have discussed earlier, the B\'{e}zier curves are transformed by projective transformations which have 8 degrees of freedom (dof) with greater flexibility, and the primitive shapes are transformed via affine transformation which has 6 dof and more control points than the B\'{e}zier curves which jointly restrict their transformation flexibility. This experiment aims to explain the effect of projective transformations on primitive shapes and why we cannot use them during canvas completion with primitive shapes. \cref{fig:ab_trans} shows the effect of affine as well as projective transformations on a simple trapezoidal like shape %\anjan{the figure does not look like house. I would call it as a generic shape or trapezoidal may be}
made of simple primitives.

It has been observed that, after an affine transformation, shapes are only rotated, scaled, or translated however, they maintain their shape characteristics. This is because an affine transformation is defined by a $3 \times 3$ matrix of the following form: A =
$\begin{bmatrix}
A_{11} & A_{12} & t_x\\
A_{21} & A_{22} & t_y\\
0 & 0 & 1
\end{bmatrix}$ = 
$\begin{bmatrix}
A & t \\
0^T & 1
\end{bmatrix}$, where $t = (t_x, t_y)$ is the translation vector comprising the amount of translation along the x-axis and y-axis, respectively. Now, in order to understand the transformations provided by the matrix $2 \times 2$ square matrix $A$, we perform a singular value decomposition (SVD) of $A$ as follows:
\begin{equation}
\resizebox{0.90\columnwidth}{!}{$
    A \xrightarrow[]{\text{SVD}} USV^T = (UV^T)(VSV^T) = R(\theta)R(-\phi)SR(\phi)$}
\end{equation}
where $UV^T = R(\theta)$ and $V^T = R(\phi)$. This SVD always provides two orthonormal matrices $U$ and $V$ and a diagonal matrix $S$ containing the eigenvalues of the matrix $A$. As the rotation matrix is also orthonormal, these two orthogonal matrices depict two rotations and provide 2 degrees of freedom. on the other hand, S is a diagonal matrix containing 2 eigenvalues of $A$ (As $A$ is a $2 \times 2$ matrix); each eigenvalue represents one scaling factor providing 1+1 = 2 degrees of freedom. As we do not get any morphing operation via SVD decomposition, the characteristics of the shape will be preserved after deformation, which helps the player to identify it in the final canvas.

Similarly, projective transformations are also performed with a $3 \times 3$ matrix of the form $P =$ 
$\begin{bmatrix}
A_{11} & A_{12} & t_x\\
A_{21} & A_{22} & t_y\\
v_{11} & v_{12} & v_{13}
\end{bmatrix}$ = 
$\begin{bmatrix}
A & t\\
v^T & v
\end{bmatrix}$, where $v = (v_{11}, v_{12}, v_{13})$ controls the shape deformation. The last row of $P$ is not $[0, 0, 1]$, which causes the shape deformation. We cannot use this projective transformation on primitive shapes as in Pictionary games, where one player builds an object part by part, and the other has to recognize the parts. If the shapes get deformed, they lose their recognizability, which motivates us to restrict transformations to an affine transformation.

\section{Primitive based canvas completion}
\label{supp:04}
\begin{figure*}[!ht]
\centering
  \includegraphics[width=0.95\textwidth]{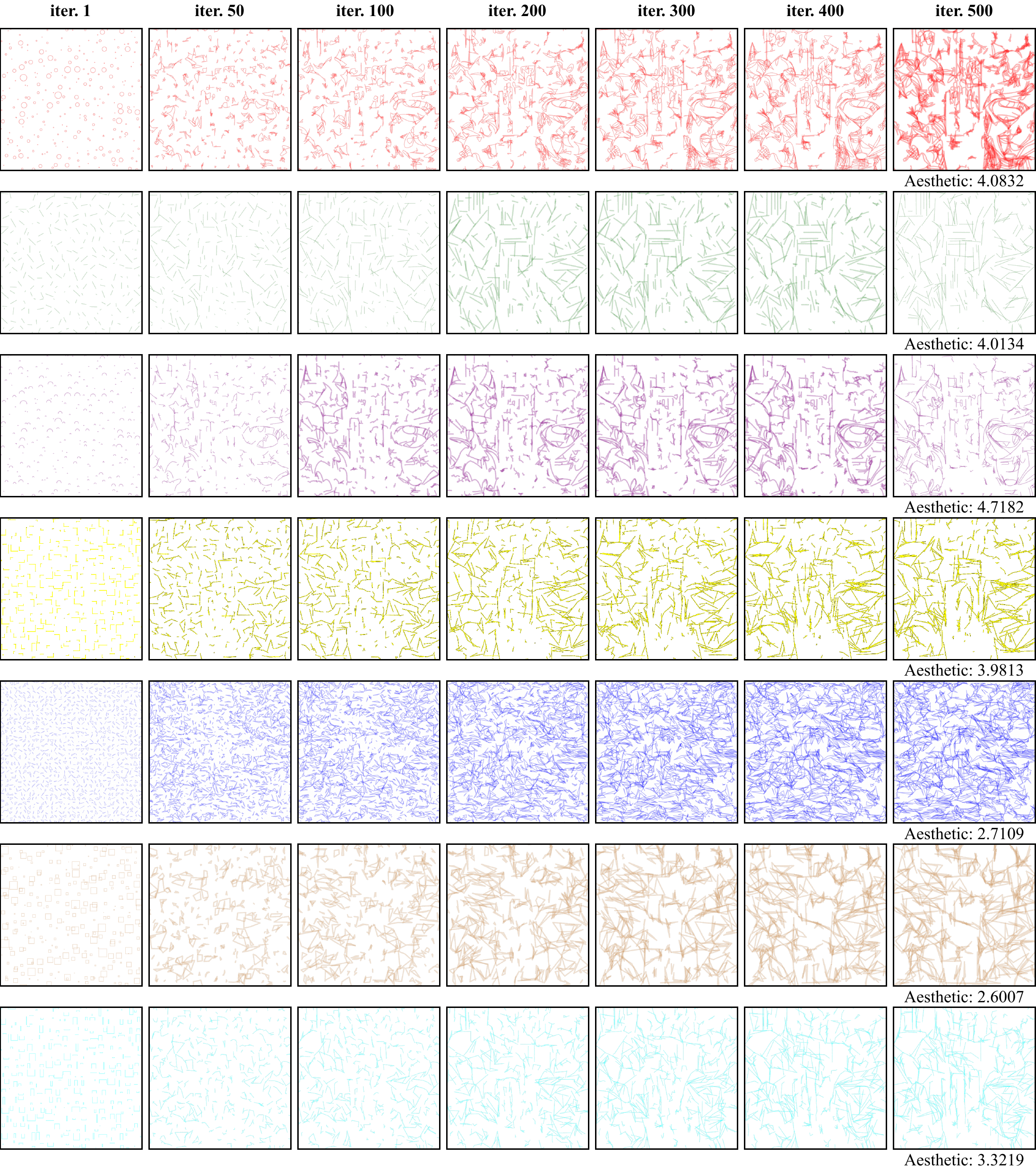}
%\Description[]{}
%\vspace{-8mm}
\caption{\textbf{Evolution of the primitive shapes:} \textcolor{red}{circle}, \textcolor{green}{line}, \textcolor{purple}{semi-circle}, \textcolor{yellow}{L-shape}, \textcolor{blue}{triangle}, \textcolor{brown}{square}, and \textcolor{cyan}{U-shape} with random initialization for the prompt "Batman and Superman, with smoke and explosions in the background". Over the iteration \textcolor{purple}{semi-circle} generates the best quality SVGs compared to the rest.}
\label{fig:ab_shape_evol}  % Give a unique label
\vspace{-4mm}
\end{figure*}

\begin{figure*}[!ht]
\centering
  \includegraphics[width=0.95\textwidth]{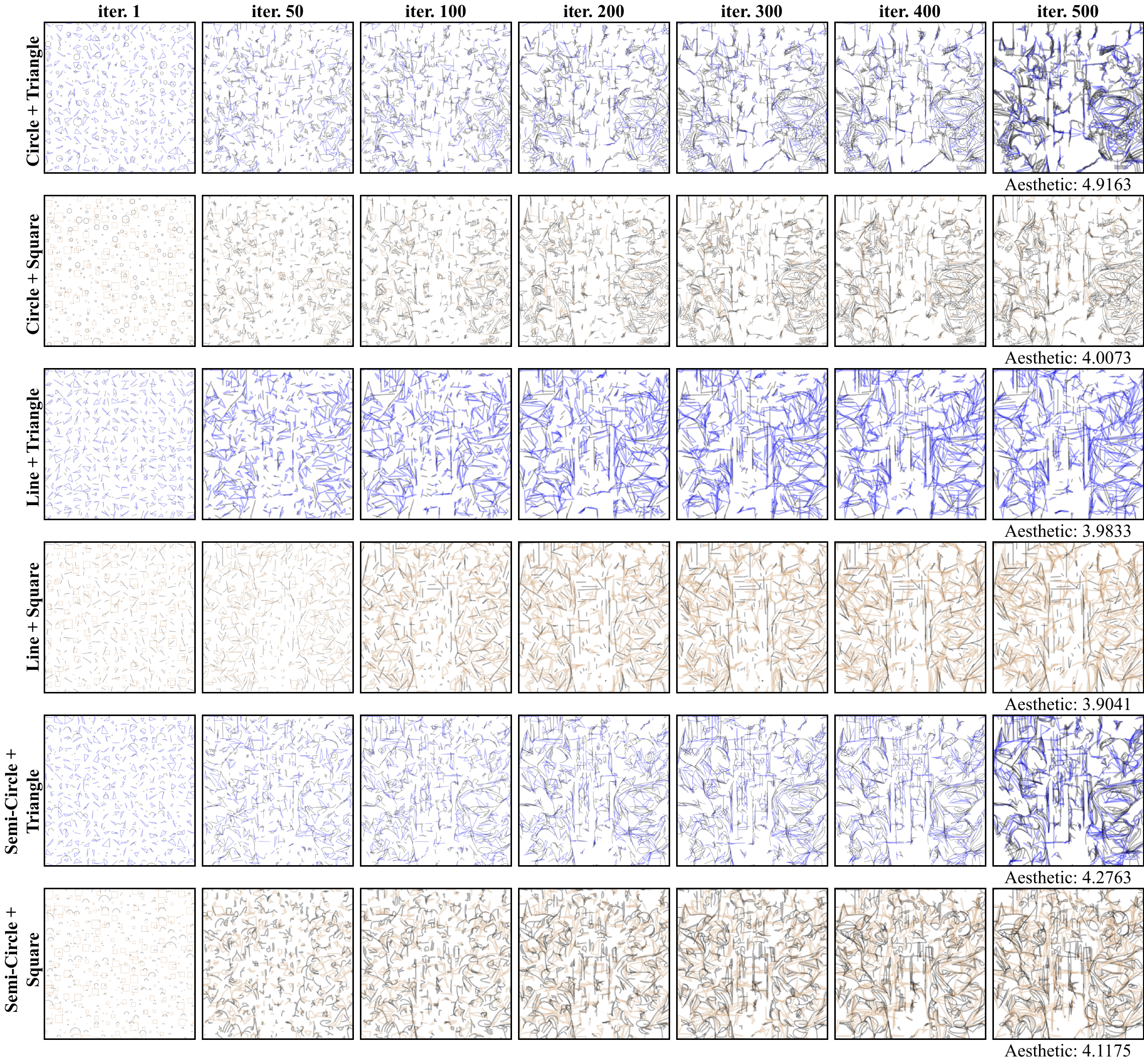}
%\Description[]{}
\caption{\textbf{Mixing of primitive shapes:} We have used circles, semi-circles, and lines as our base shape with triangles and squares as an auxiliary one. Although it improves the aesthetic score for circle+triangle, it provides messier SVGs for the rest. }
\label{fig:ab_shape_evol_two}       % Give a unique label
%\vspace{-4mm}
\end{figure*}

\myparagraph{Problem Statement:}
Primitive-based canvas completion performs the same synthesis through optimization in a more constrained environment. A primitive shape is a special form of B\'{e}zier curves $\{S_i\}_{i=1}^n = \{s_i,w_i, \gamma_i \}_{i=1}^n$, where $S_i = \{s_i\}_{j=1}^p = \{x_i, y_i\}_{j=1}^p; \forall p\in\{2,3,4\}$ and they can only go through the affine transformation (6 dof: rotation, translation and anisotropic-scaling) during optimization with $\mathcal{R}_d$. This effect of transformation constraints has been provided in \cref{fig:exp1}, \cref{tab:02} and \cref{fig:ab_shape_evol}, \cref{fig:ab_shape_evol_two}.

As we proposed primitive-based scene completion as our downstream task, we decided to study the evolution of the primitive shapes without per-box mask latent base canvas initialization (i.e., without semantic guidance). This experiment aims to understand the true potential of primitive shapes toward canvas completion in a restricted environment with minimal guidance or no guidance at all. We have decided to initialize the canvas with primitive shapes in a random manner, as semantic guidance provides a strong initialization and shape guidance during iterative evolution, which significantly forces the shape evolution to maintain the semantic pattern. The results we have obtained via this experiment have been depicted in \cref{fig:ab_shape_evol}.

The semi-circle obtained the most aesthetically pleasing SVG (mean Aesthetic score: 5.0241), maintaining the text prompt due to open-ended and fewer control points (only 2), consolidating the claim we have obtained in \ref{tab:02} of the main paper. Similarly, the square shape provides the worst result due to its closed nature and higher number of control points (i.e., 4), constraining its transformation flexibility. However, we have an important observation in circle evolution, which produces a messier SVG compared to the one with semantic guidance. As the circle has a closed-form shape with only 2 control points, it has significantly less flexibility than a semicircle during evolution. So, during optimization without semantic guidance, it is unable to optimize the unwanted shapes initialized randomly.

However, from the \cref{tab:02}, and \cref{fig:ab_shape_evol} of this supplementary, it can be concluded that circle, semi-circle, and lines are the best possible shapes to complete a canvas, and square and triangle are the worst ones. However, we are keen to know if the shapes with a higher number of control points (square, triangle, and so on) can help the other shapes (circle, semi-circle, and line) in an auxiliary manner to generate more aesthetically promising SVGs. The outcomes of this study have been depicted in \cref{fig:ab_shape_evol_two}.

As shown in \cref{fig:ab_shape_evol_two}, the combination of circle+triangle improves the aesthetic score (4.0832 to 4.9163) from the SVG generated with only circles. Not only that, we can observe a certain failure for the rest of the combinations as the synthesized SVG contains a lot of unnecessary strokes that are not properly optimized during evolution. As circles and triangles have more inherent geometric compatibility compared to other combinations, the circular shape can easily accommodate the angular structure of a triangle without any overlapping or awkward intersections. Other combinations, like line and square, might not fit together as neatly due to differences in shape and curvature. Also, the circle-triangle combination creates a perception of symmetry or balance that the other combinations lack. Symmetry is often considered aesthetically pleasing, and the circular shape paired with the triangular shape creates a sense of balance that is missing in the other combinations.

\begin{table*}[!htbp]
%\vspace{-4mm}
\centering
\caption{Evaluation of canvas completions with primitive shapes.}
\resizebox{0.7\textwidth}{!}{
\begin{tabular}{c|c|c|c|c|c|c|c|c}
\hline
Shapes/Metric &
  \begin{tabular}[c]{@{}c@{}}CS $\uparrow$ \cite{huang2024t2i}\end{tabular} &
  \begin{tabular}[c]{@{}c@{}}FID $\downarrow$ \cite{heusel2017gans}\end{tabular} &
  \begin{tabular}[c]{@{}c@{}}PSNR $\uparrow$ \cite{hore2010image}\end{tabular} &
  \begin{tabular}[c]{@{}c@{}}CLIP-T $\uparrow$ \cite{radford2021learning}\end{tabular} &
  \begin{tabular}[c]{@{}c@{}}BLIP $\uparrow$ \cite{li2022blip}\end{tabular} &
  \begin{tabular}[c]{@{}c@{}}Aes. $\uparrow$ \cite{Aes2022}\end{tabular} &
  \begin{tabular}[c]{@{}c@{}}HPS $\uparrow$ \cite{wu2023human}\end{tabular} &
  \begin{tabular}[c]{@{}c@{}}Conf. $\uparrow$ \cite{xing2023diffsketcher}\end{tabular} \\ \hline
Circle      & 0.4817          & 76.38          & 10.24          & 0.2974          & 0.4265          & 4.7531          & 0.2231          & 0.39          \\
Line        & 0.4566          & 82.37          & 7.62           & 0.2369          & 0.3664          & 4.1522          & 0.1976          & 0.32          \\
Semi-Circle & \textbf{0.5194} & \textbf{61.34} & \textbf{12.22} & \textbf{0.3145} & \textbf{0.4578} & \textbf{5.1362} & \textbf{0.2413} & \textbf{0.47} \\
Triangle    & 0.2237          & 111.54         & 2.11           & 0.1433          & 0.2237          & 3.0567          & 0.1032          & 0.24          \\
Square      & 0.2013          & 127.83         & 1.07           & 0.1123          & 0.1952          & 2.3645          & 0.0914          & 0.11          \\
L-shape     & 0.3176          & 94.56          & 5.67           & 0.1958          & 0.2987          & 3.7134          & 0.1532          & 0.29          \\
U-shape     & 0.3124          & 96.82          & 4.12           & 0.1663          & 0.2637          & 3.228           & 0.1338          & 0.24          \\ \hline
\end{tabular}
}
\label{tab:02}
\vspace{-4mm}
\end{table*}

\subsection{Details of user study}
\label{supp:08}
We have conducted the perceptual study in order to evaluate the credibility and widespread adoption of the synthesized SVGs. We chose 60 SVGs synthesized by CLIPDraw (b/w and color) \cite{frans2022clipdraw}, CLIPDrawX \cite{mathur2023clipdrawxprimitivebasedexplanationstext}, DiffSketcher \cite{xing2023diffsketcher}, VectorFusion \cite{jain2023vectorfusion}, SVGDreamcer \cite{xing2023svgdreamer}, and CraftSVG (abstract, sketch, primitives, and CLIPArt), each style contains 10 examples. So we provide the users with 10 sets, each set containing 13 images, followed by the aforementioned style, and ask them to rate the SVGs on the scale of 1 to 5 (1 Low; 5 high) based on aesthetics, simplicity, and recognizability.
%answer the following 4 questions for each set: 1) ``Which of the SVGs is more aesthetically promising?''; 2) ``Which of the graphics are close to freehand drawing?''; 3) ``Which one do you think is AI generated?''; and lastly 4) ``Which of the graphics correctly explains the given text prompt?''. 
%Some 47 users participated in this human study. The outcome of this study has been reported in \cref{fig:users}\footnote[2]{users are allowed to choose at most 2 options and at least one style}.
\begin{figure*}[!ht]
%\vspace{-4mm}
\centering
  \includegraphics[width=\textwidth]{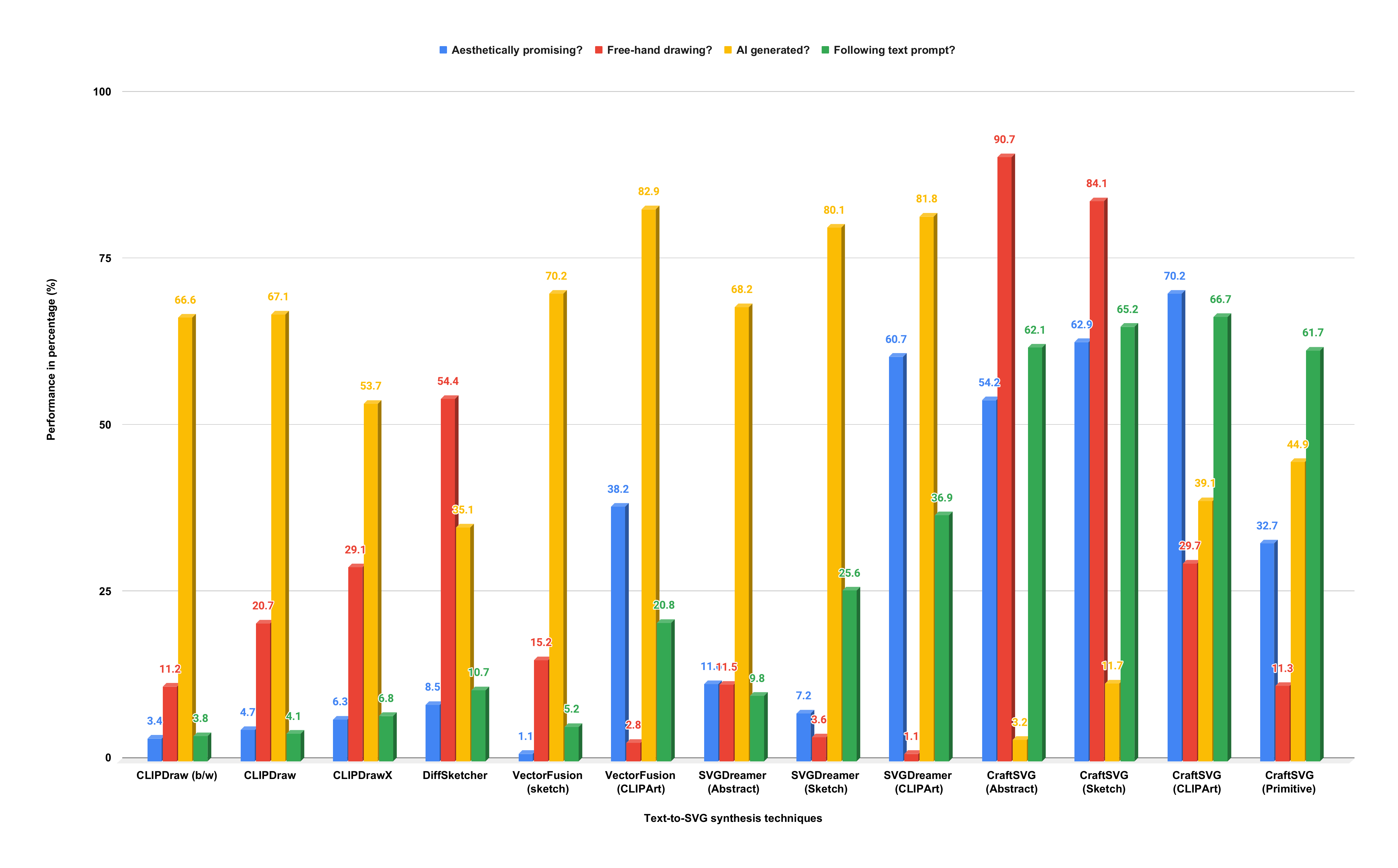}
%\Description[]{}
%\vspace{-8mm}
\caption{\textbf{User study:} CraftSVG gets the maximum vote for being aesthetically promising, whereas, VectorFusion is identified as AI generated due to excessive smoothing and color over-saturation.}
\label{fig:users}       % Give a unique label
%\vspace{-4mm}
\end{figure*}

For the user, color is proportional to the aesthetically promising; that's why CraftSVG (CLIPArt) and SVGDreamer hold the 1st and 2nd positions, respectively. Similarly, they consider simple sketches as freehand drawings, which makes CraftSVG (abstract) the winner of this category, and CraftSVG (sketch) stands second. However, the 3rd question helps us to get the confusion score (i.e., 100 - \% of people identify the image as AI generated). CraftSVG (abstract) is a clear winner, and CraftSVG (sketch) stands second in this category as the light-shedding gives an artistic touch to the synthesized SVGs. All the users agreed that the SVGs synthesized by CraftSVG properly followed the complex text prompt with enumeration and spatial relationship, which consolidated our claim in this paper.

\section{Importance of Alignment Loss}
\begin{figure}[!htbp]
    \centering
    \includegraphics[width=\columnwidth]{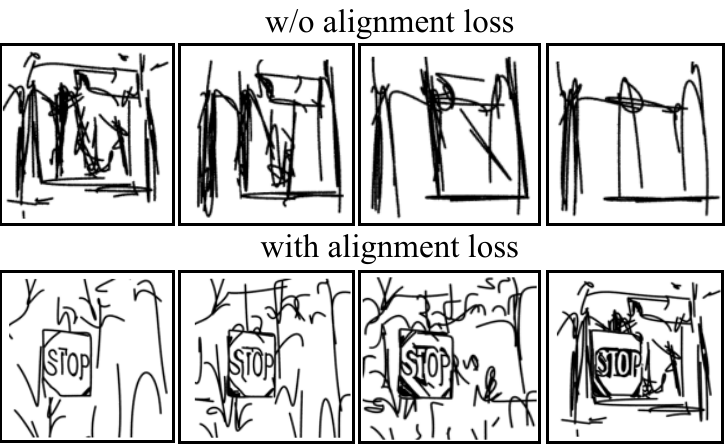}
    %\Description[]{}
    \caption{The ablation study of the alignment loss with the text prompt “The stop sign behind the fence”}
    \label{fig:abalign}
\end{figure}

In this section, we have performed the ablation study of the CraftSVG with and without using the perceptual alignment loss defined in the \cref{eq:align}. It can be done by setting the $\lambda_{align} = 0$ while optimizing the CraftSVG with $\mathcal{L}_{synth}$ as defined in \cref{eq:06}. The result of this ablation study has been reported in \cref{fig:abalign} and \cref{tab:align}, respectively.

From \cref{fig:abalign}, it can be observed that with perceptual alignment loss, the background and foreground strokes often collide with the foreground strokes, leading to noisy abstract VectorArts that do not even follow the text prompt. On the other hand, with perceptual alignment loss, CraftSVG always preserves the foreground object structure while optimizing the background strokes during alignment. Also, \cref{tab:align} depicts a significant performance improvement of CraftSVG while using the perceptual alignment loss. Also, it can help to optimize the SVG parameters along with the MLP parameters jointly to abstractify the VectorArt by maintaining enumeration, spatial arrangement, and relationship provided by a text prompt for complex scenes.

\begin{table}[!ht]
\vspace{-2mm}
\centering
\caption{Ablation of the Alignment Loss}
\label{tab:align}
\resizebox{\columnwidth}{!}{
\begin{tabular}{@{}cccccccc@{}}
\toprule
Methods/Metrics     & CS     & FID    & PSNR  & CLIP-T & BLIP-T & Aes    & HPS    \\ \midrule
w/o alignment loss  & 0.2181 & 122.62 & 4.33  & 0.1484 & 0.2575 & 3.2863 & 0.1877 \\
With alignment loss & 0.6176 & 51.42  & 15.98 & 0.4563 & 0.5223 & 5.9873 & 0.3167 \\ \bottomrule
\end{tabular}
}
%\vspace{-4mm}
\end{table}

\section{Attention-based initialization comparison}
\begin{figure}[!htbp]
    \centering
    \includegraphics[width=\columnwidth]{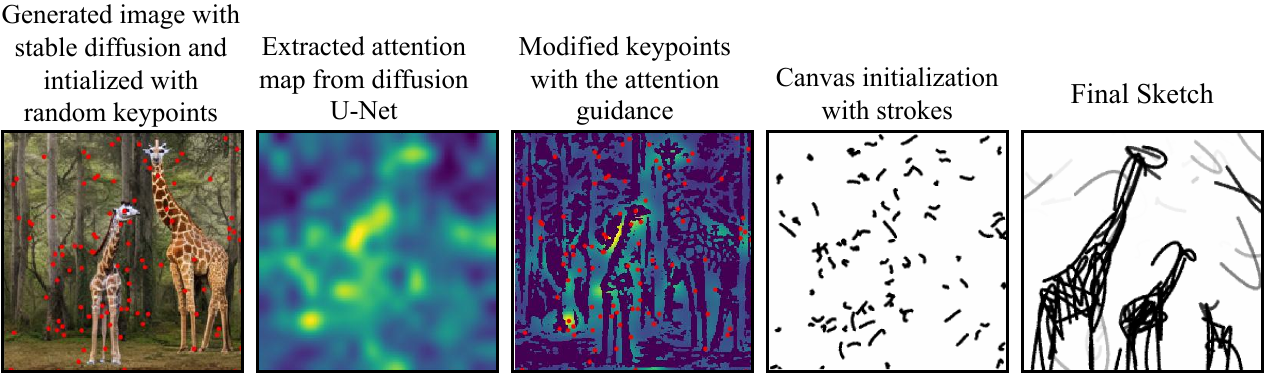}
    %\Description[]{}
    \caption{Attention-based initialization used in the DiffSketcher \cite{xing2023diffsketcher}. For the prompt "Two giraffes and three elephants", it directly initializes the keypoints from the attention map diffusion U-Net.}
    \label{fig:diffab}
\end{figure}

\begin{figure}[!htbp]
    \centering
    \includegraphics[width=\columnwidth]{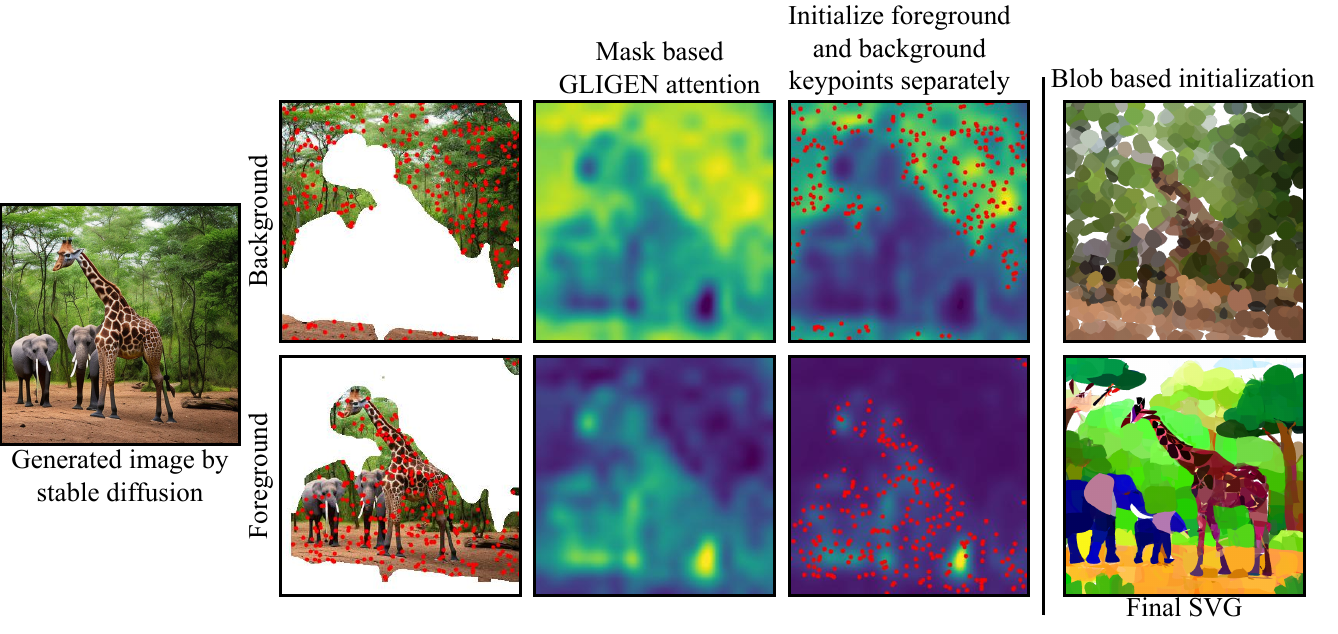}
    %\Description[]{}
    \caption{Attention-based initialization used in the SVGDreamer \cite{xing2023svgdreamer}. For the prompt "Two giraffes and three elephants in a forest", it creates the foreground mask to separate the objects from the background and initialize blobs of those two regions separately.}
    \label{fig:svdab}
\end{figure}

\begin{figure}[!htbp]
    \centering
    \includegraphics[width=\columnwidth]{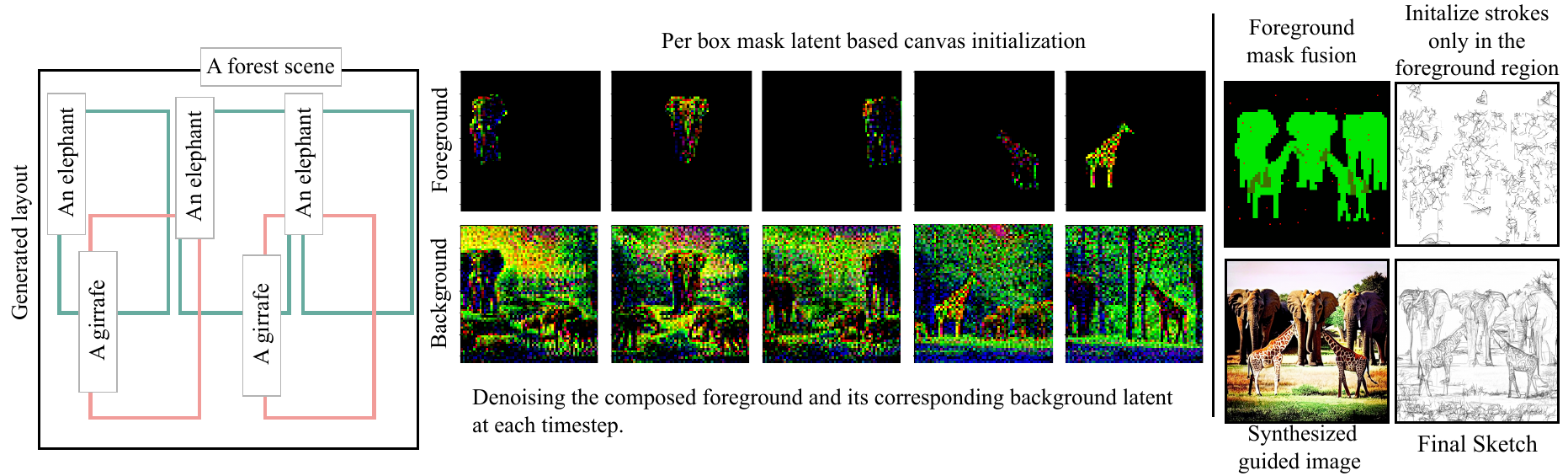}
    %\Description[]{}
    \caption{Attention-based initialization used in CraftSVG. For the prompt "Two giraffes and three elephants in a forest", it follows a progressive approach via layout guidance and initializes strokes only in the foreground region, and optimizes stroke length along with its color and width to complete the canvas.}
    \label{fig:craftsvgab}
\end{figure}

All the diffusion-based text-to-SVG synthesis techniques first synthesize a guided image from the text prompt and then optimize the strokes/blobs in the canvas by maximizing the similarity with the guided image. It has been interesting to study how attention-based initialization makes a difference in the final SVG and helps in faster convergence (see \cref{fig:diffab}, \cref{fig:svdab}, \cref{fig:craftsvgab}). Diffsketcher \cite{xing2023diffsketcher} utilizes the diffusion U-Net map on the synthesized image to identify the keypoints for stroke initialization. Although it helps in faster convergence, it is unable to follow the enumeration and spatial relationship mentioned in the input text prompt. Similarly, SVGDreamer creates separate semantic maps for foreground and background objects using the text encoder's cross-attention map on the synthesized guided image, optimizing them independently with attention mask-based loss. This reduces interaction between elements, making the drawing appear AI-generated. It has to be noted that both methods first create the synthesized guided image and then try to get the attention map, which is unable to control the object counting and their position in the canvas. On the contrary, CraftSVG first generates a layout from the input text prompt via LLM in order to follow the enumeration and spatial relationship mentioned in the prompt. Then, we use per-box mask latent guidance, processing a single foreground object box through the diffusion U-Net to extract coherent foreground and background masks. By iteratively merging these masks, CraftSVG maintains depth information, allowing opacity control, where strokes of closer objects have higher opacity. Also, we initialize strokes only in the foreground region and then optimize the length of the strokes along with their width and color by maximizing the similarity with the synthesized guided image obtained through layout guidance.

\section{Comparison with T2I diffusion models}
\label{supp:09}
\begin{figure}[!ht]
\centering
  \includegraphics[width=\columnwidth]{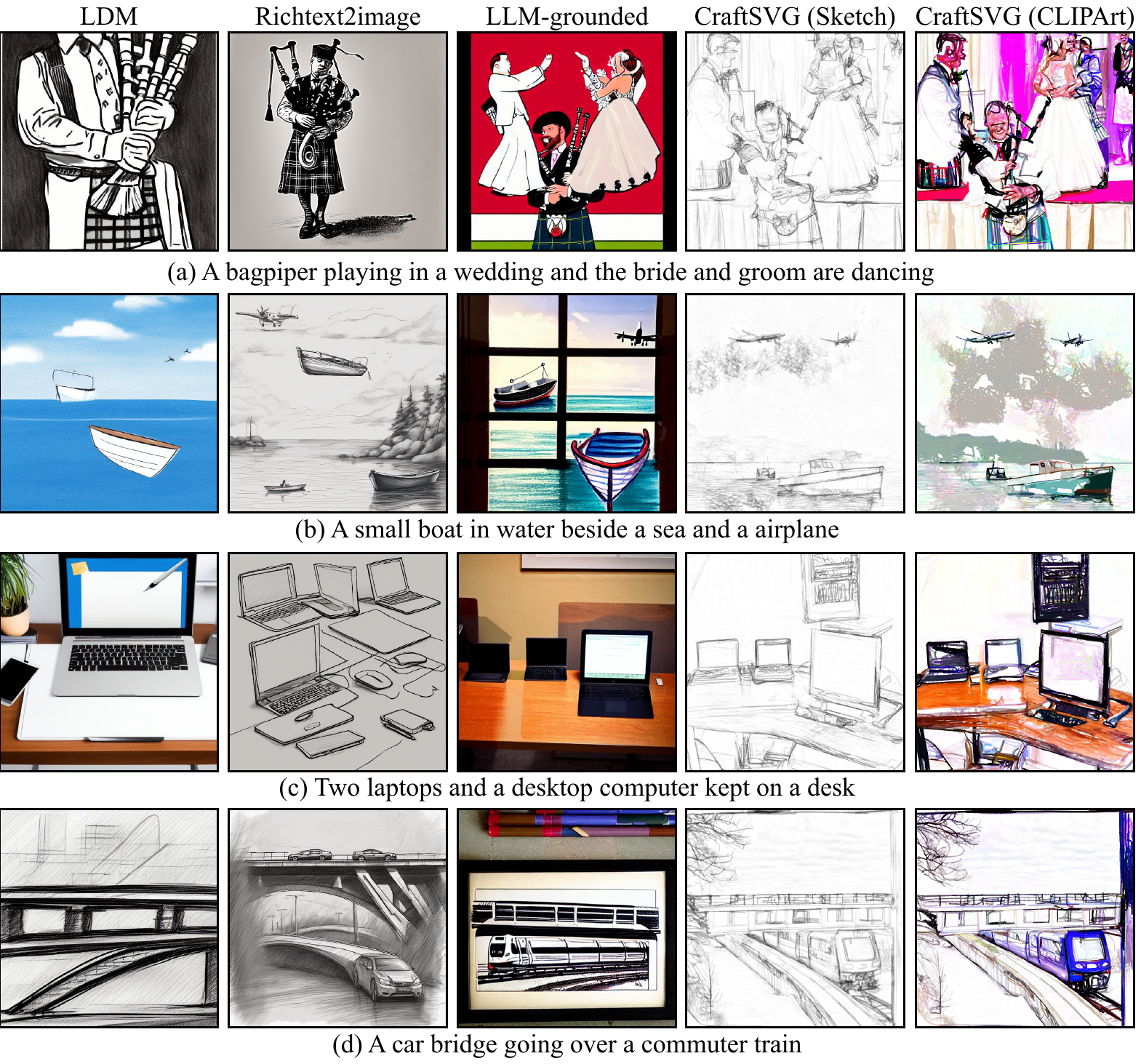}
%\Description[]{}
%\vspace{-6mm}
\caption{Comparison with text-to-image synthesis techniques.}
%\vspace{-6mm}
\label{fig:comp_diff}       % Give a unique label
%\vspace{-6mm}
\end{figure}

With the recent advent of text-to-image diffusion models \cite{schaldenbrand2022styleclipdraw,lian2023llm}, \cite{ge2023expressive} we are keen to know about their potential to replicate freehand drawing style. In order to perform this evaluation we compare the SVGs synthesized by CraftSVG with the image generated by the diffusion models. In order to encourage the results to be abstract and follow the free-hand sketch style, we append a suffix to
the text prompt: ``A free-hand sketch of xxx'' as mentioned in \cite{xing2023diffsketcher}. The obtained qualitative as well as quantitative results are reported in \cref{fig:comp_diff} and \cref{supp_tab:01} respectively. As depicted in \cref{fig:comp_diff}, LDM \cite{Rombach2022LatentDiffusion} and Richtext2image \cite{ge2023expressive} are unable to follow the proper enumeration and spatial relationship mentioned in the text prompt. In \cref{fig:comp_diff}(c), LDM generated a real image, and in \cref{fig:comp_diff}(b), Richtext2image draws a boat in the sky, which is beyond specification. Although LLM-grounded diffusion \cite{lian2023llm} follows the text prompt, it is unable to make the freehand sketch effect showing the importance as well as the superiority of our proposed CraftSVG. It not only follows the enumeration and spatial relationship but also creates the specific style mentioned by the user.

\begin{table*}[!htbp]
\centering
\caption{Quantitative evaluation on conventional text-to-image synthesis techniques.}
%\vspace{-4mm}
\resizebox{0.8\textwidth}{!}{
\begin{tabular}{l|c|c|c|c|c|c|c}
\hline
Method / Metric &
  \begin{tabular}[c]{@{}c@{}}CS $\uparrow$ \cite{huang2024t2i}\end{tabular} &
  \begin{tabular}[c]{@{}c@{}}FID $\downarrow$ \cite{heusel2017gans}\end{tabular} &
  \begin{tabular}[c]{@{}c@{}}PSNR $\uparrow$ \cite{hore2010image}\end{tabular} &
  \begin{tabular}[c]{@{}c@{}}CLIP-T $\uparrow$ \cite{radford2021learning}\end{tabular} &
  \begin{tabular}[c]{@{}c@{}}BLIP $\uparrow$ \cite{li2022blip}\end{tabular} &
  \begin{tabular}[c]{@{}c@{}}Aes. $\uparrow$ \cite{Aes2022}\end{tabular} &
  \begin{tabular}[c]{@{}c@{}}HPS $\uparrow$ \cite{wu2023human}\end{tabular} \\ \hline
\begin{tabular}[c]{@{}l@{}}LDM  \cite{rombach2022high}\end{tabular} &
  0.4906 &
  99.18 &
  12.24 &
  0.3512 &
  0.4221 &
  4.9317 &
  0.2332 \\ \hline
\begin{tabular}[c]{@{}l@{}}Richtext2image  \cite{ge2023expressive}\end{tabular} &
  0.4130 &
  87.04 &
  14.34 &
  0.3772 &
  0.4017 &
  \textbf{7.6468} &
  0.2661 \\ \hline
\begin{tabular}[c]{@{}l@{}}LLM-grounded diffusion  \cite{lian2023llm}\end{tabular} &
  0.5052 &
  69.18 &
  15.08 &
  0.3815 &
  0.4327 &
  5.9821 &
  0.2858 \\ \hline
CraftSVG (Sketch) &
  0.6342 &
  48.42 &
  16.07 &
  0.4563 &
  0.5223 &
  6.7832 &
  0.3167 \\
CraftSVG (CLIPArt) &
  \textbf{0.7091} &
  \textbf{39.87} &
  \textbf{17.15} &
  \textbf{0.5013} &
  \textbf{0.5783} &
  7.0779 &
  \textbf{0.3523} \\ \hline
\end{tabular}
}
\label{supp_tab:01}
\vspace{-4mm}
\end{table*}

Similarly, we have observed an interesting fact in \cref{supp_tab:01}. The aesthetics of Richtext2image is slightly higher than the proposed CraftSVG as it generates high-quality pixel graphics ($1024 \times 1024$), whereas LDM, LLM-grounded diffusion, as well as CraftSVG work on the canvas size of $512 \times 512$. Although Richtext2image generates high-quality images, its background exhibits an old paper texture, which looks like an old grayscale photograph rather than sketches. This comes from the bias of the training dataset used by Richtext2image.  Conversely, the sketches generated by our CraftSVG bear a closer resemblance to free-hand sketching styles and display varying degrees of abstraction.

\begin{figure}[!ht]
\centering
  \includegraphics[width=\columnwidth]{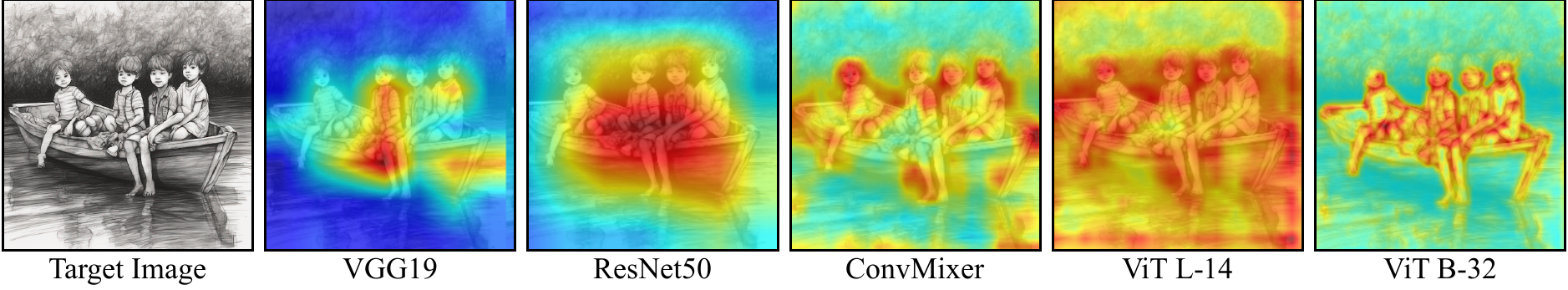}
%\Description[]{}
\caption{Comparison of different types of image encoders based on their feature extraction performance.}
\label{fig:enc}       % Give a unique label
\vspace{-4mm}
\end{figure}

\section{Ablation of the image encoders}
\label{supp:new2}
As we perform similarity maximization between the encoded features of the rasterized target image $\mathcal{I}_r$ and the differential renderer $\mathcal{R}_d (\theta)$, we should stress about choosing the best feature extractor. However, there are no such metrics to compare the feature extraction performance between different backbones. In order to compare the different image encoders $\mathcal{I}$, we have tried to extract their activation map/ attention map of the last layers for Convolution/transformer-based architectures. The obtained results have been reported in \cref{fig:enc}. As we can observe in \cref{fig:enc}, the Grad-CAM results of the last layer of VGG19 don't capture all the required features to complete the scene, leading to the synthesized incomplete sketches. With ResNet-50, we can improve the performance, however, it only focuses on the main objects, not on the background scenarios. Moreover, with the attention used in the ConVMixer model \cite{trockman2022patches}s, we can capture the global perspective but miss the local minute details. So we have decided to use the CLIP ViT B-32, which divides the input image into patches to get the global perspective and uses transformer layers to capture the local one, leading to use this as our desired image encoder in CraftSVG.
\section{Effect of bounding box scaling}
\label{supp:05}
\begin{figure*}[h!]
\centering
  \includegraphics[width=0.9\textwidth]{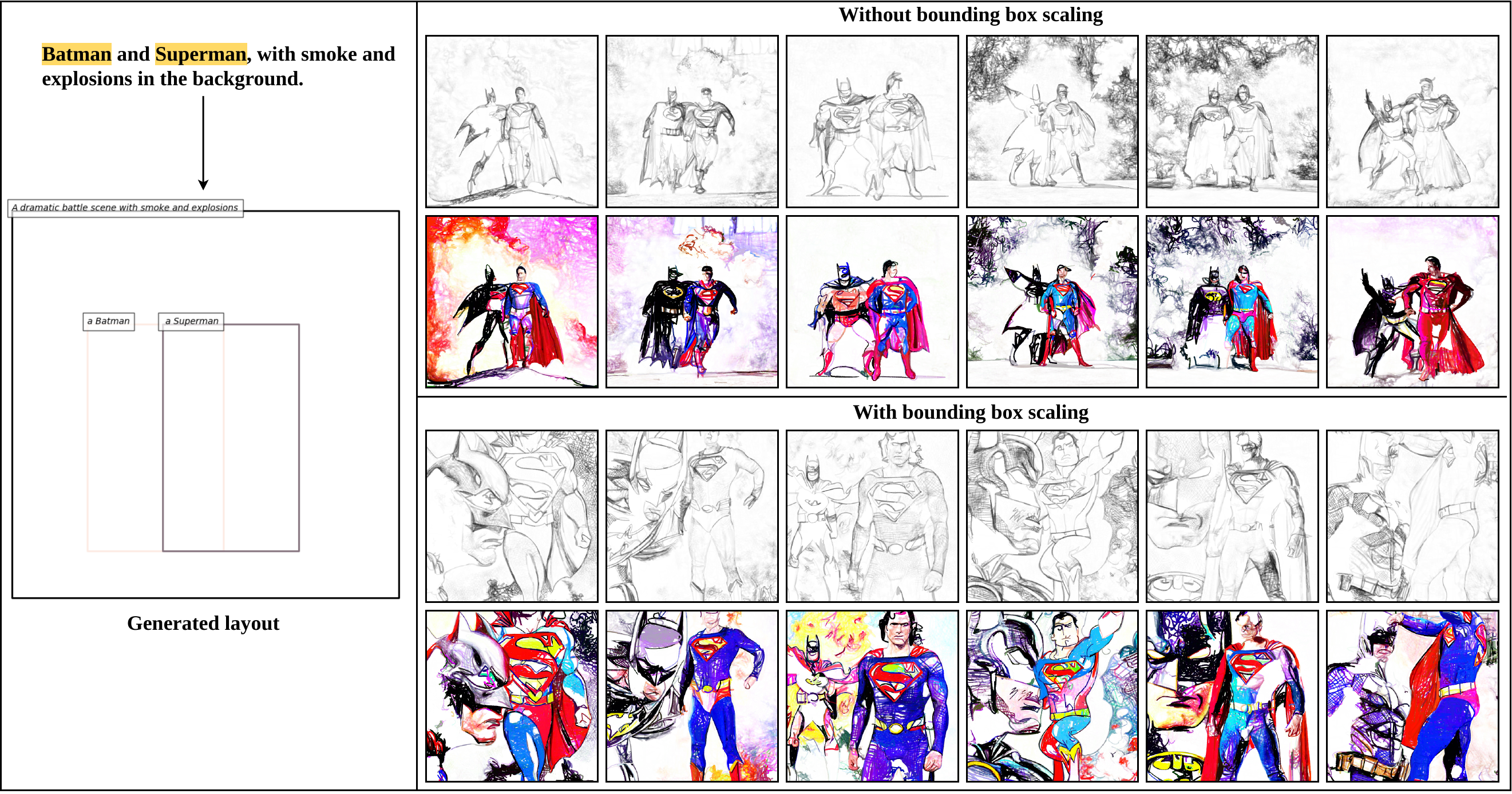}
%\Description[]{}
\caption{\textbf{Effect of bounding box scaling:} it enables the CraftSVG to create more creative, visually appealing SVGs, maintaining the contextual information provided in the text prompt by relaxing the constrained area specified in a bounding box.}
\label{fig:bb_scale}       % Give a unique label
\end{figure*}

Vector graphics synthesis operates solely within the bounding box, often producing results that are overly constrained and lack contextual coherence beyond the specified boundaries. This approach may lead to incomplete or fragmented images that fail to capture the broader scene or object context, limiting the model's ability to generate creative and visually appealing outputs. In order to address this issue, we have introduced bounding box scaling techniques, which enable us to use the full canvas for generating objects instead of the specified boxes. \cref{fig:bb_scale} demonstrates the SVGs synthesized with and without bounding box scaling, keeping the other attributes (text prompt, layout, no. of strokes, stroke width, opacity, and so on) unaltered.

In \cref{fig:bb_scale}, scaling enables the incorporation of surrounding elements and spatial relationships, enriching the generated images with more coherent and contextually relevant details. This broader perspective enhances the model's capacity to generate visually convincing and semantically meaningful images that better reflect objects on the canvas. However, some results can be penalized as they cut out the face in the 2nd and 6th columns of \cref{fig:bb_scale} as the amount of the scaling is controlled by the model implicitly. Still, scaling bounding boxes offers a more creative approach to vector graphics generation, facilitating the creation of more immersive and contextually rich visual outputs.

\section{Applications}
\label{supp:07}
\begin{figure*}[!htbp]
\centering
  \includegraphics[width=0.9\textwidth]{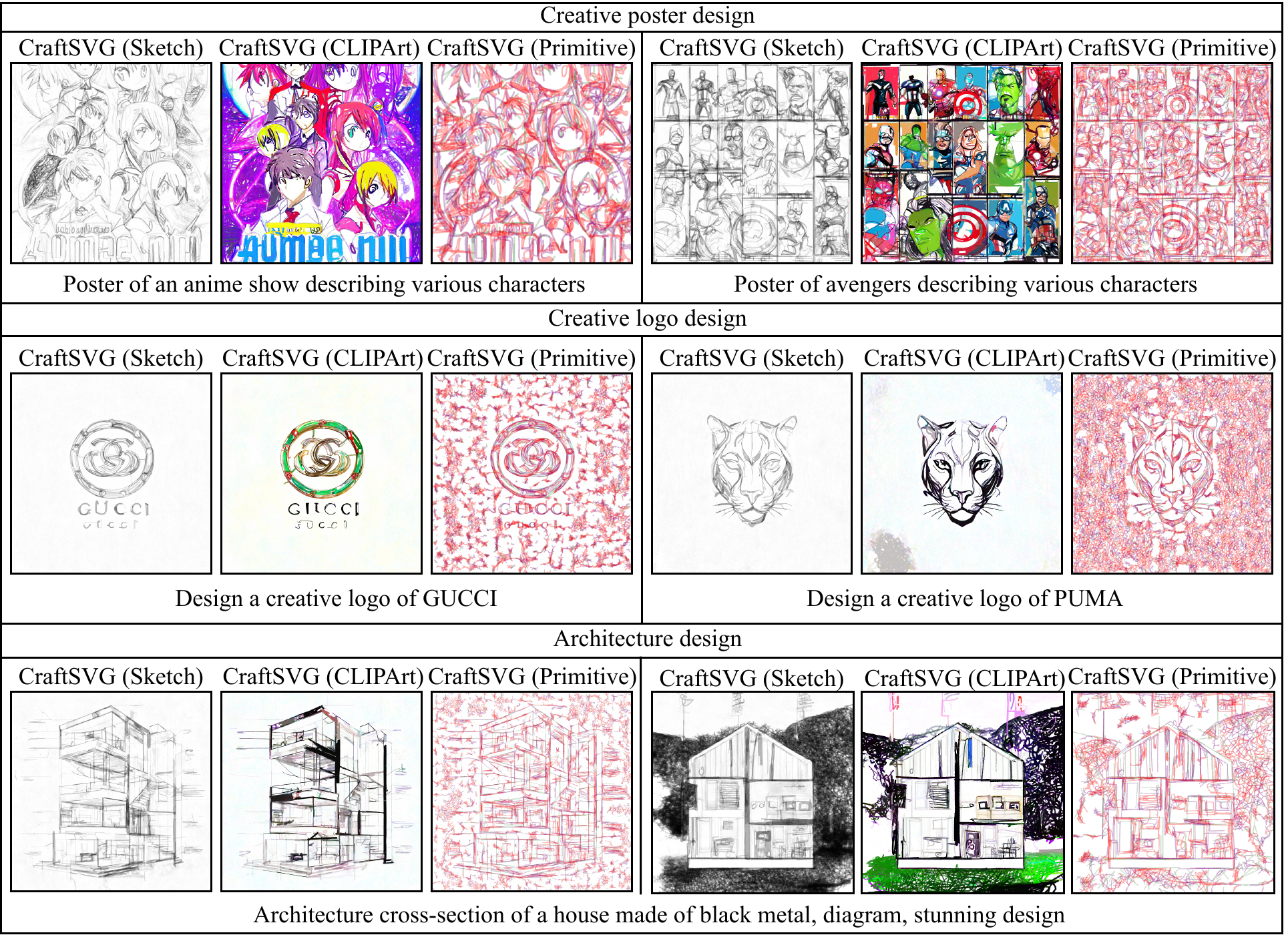}
%\Description[]{}
\caption{\textbf{Applications of CraftSVG:} It can design simple architecture besides creative posters and logos without any additional help from the text style generation module.}
\label{fig:appl}       % Give a unique label
%\vspace{-4mm}
\end{figure*}

The main application of SVG generation lies in creative logo design and creative poster design. We have also added another domain of application, namely, simple architecture design, as shown in \cref{fig:appl}. It has been observed that the proposed CraftSVG (third row of \cref{fig:appl}) can produce an effective architectural design mentioned in a text prompt.
%\vspace{-6mm}

Not only that, most of the recent work mentioned the issues of text generation in the poster as well as creative logo design \cite{rombach2022high} \cite{yang2024glyphcontrol}. In order to address this issue, SVGDreamer \cite{xing2023svgdreamer} utilized an additional text style generation module, GlyphControl \cite{yang2024glyphcontrol}, which maintains the coherency between the visual and textual content of the poster. On the other hand, CraftSVG can successfully generate visual as well as textual content without any additional help. 

\section{More qualitative examples}
\label{supp:12}
This section is dedicated to demonstrating the diverse power of the CraftSVG by generating diverse SVGs from the same as well as different text prompts. In order to enhance the readability, we have further categorized the qualitative examples in the following subsections. It has to be noted that, all the examples have been generated in the following three styles: sketch, CLIPArt, and primitive shape. As it is already proven that, circles, semi-circles, and lines generate more aesthetically promising SVGs, we use the combinations of these three types of shapes to complete the canvas. We make some exceptions to this rule in order to produce variability in the demonstration.

\begin{figure*}[!ht]
\centering
\subfloat[A dragon flying in the sky, full body]{\includegraphics[width=\textwidth]{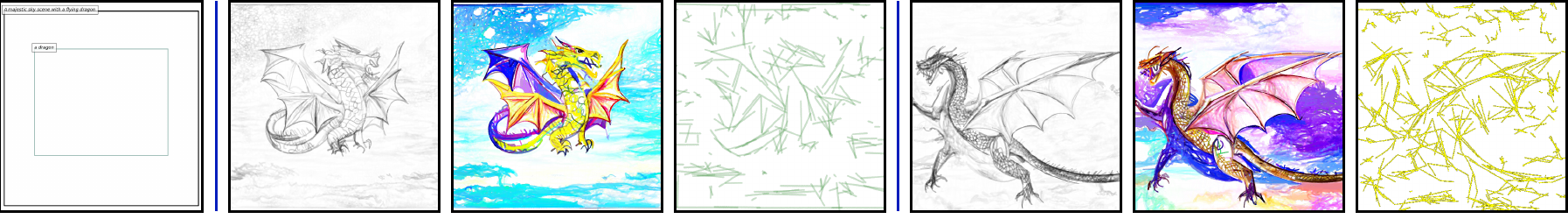}}\hfill \\
\subfloat[A unicorn is running on the grassland]{\includegraphics[width=\textwidth]{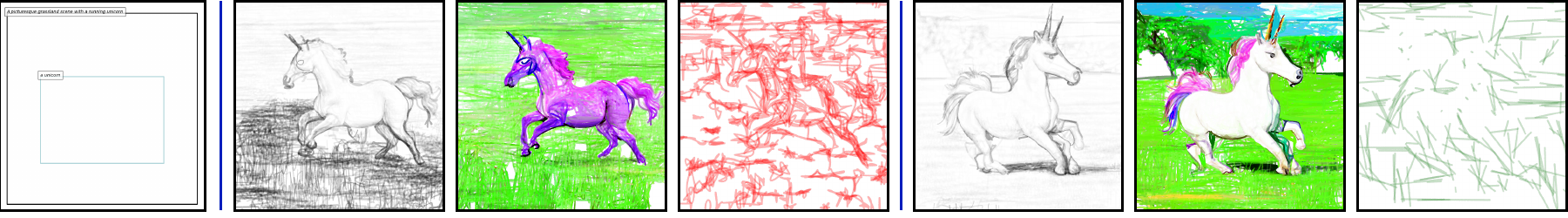}}\hfill \\
\subfloat[A yeti is taking a selfie]{\includegraphics[width=\textwidth]{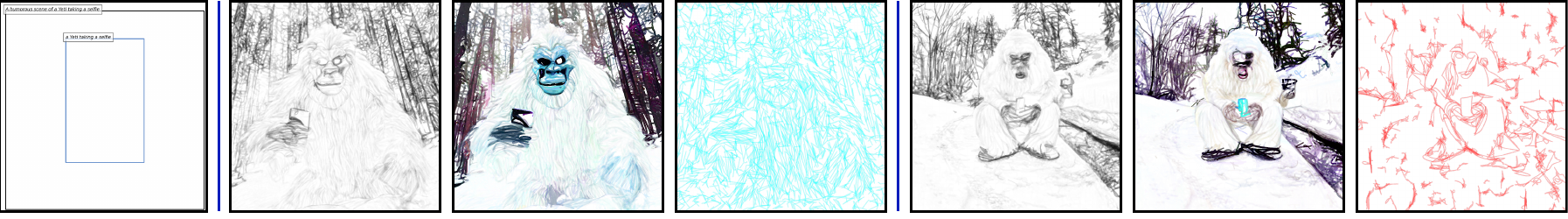}}\hfill \\
\subfloat[Very detailed masterpiece painting of baby yoda holding a lightsaber]{\includegraphics[width=\textwidth]{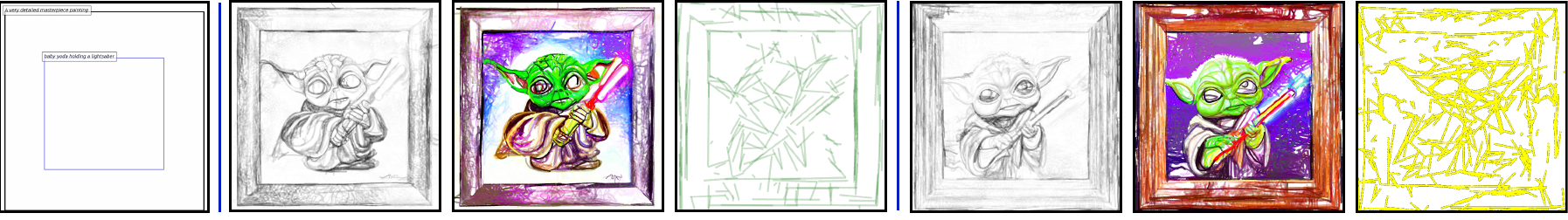}}\hfill
\caption{Non-existance object synthesis with CraftSVG}\label{fig:nonexist}
\end{figure*}

\myparagraph{Imaginary object synthesis:}
This section deals with imaginary objects (\eg, dragon, unicorn, yeti, baby Yoda, and so on) synthesized by CraftSVG. The qualitative results have been reported in \cref{fig:nonexist}. It is interesting to see the creativity of the CraftSVG to make unicorns in purple color, which is beyond imagination as in most of the existing examples, a unicorn is usually generated in white color.

\begin{figure*}[!ht]
\centering
\subfloat[A \colorbox{yellow}{horse} to the left of a \colorbox{yellow}{panda}]{\includegraphics[width=0.8\textwidth]{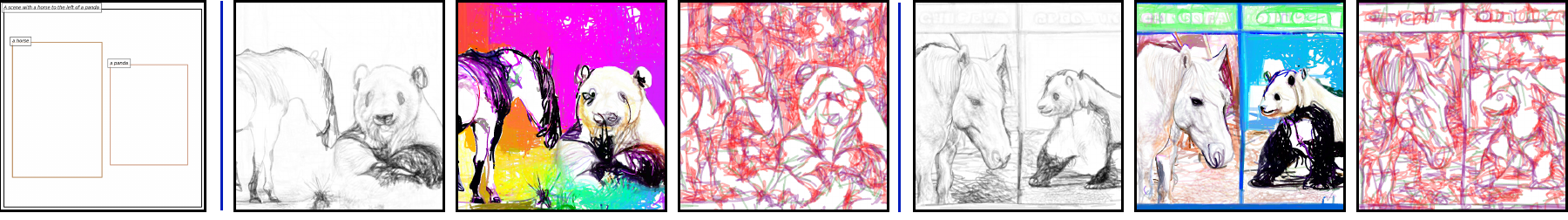}}\hfill \\
\subfloat[a \colorbox{yellow}{bowl} to the left of an \colorbox{yellow}{orange}
]{\includegraphics[width=0.8\textwidth]{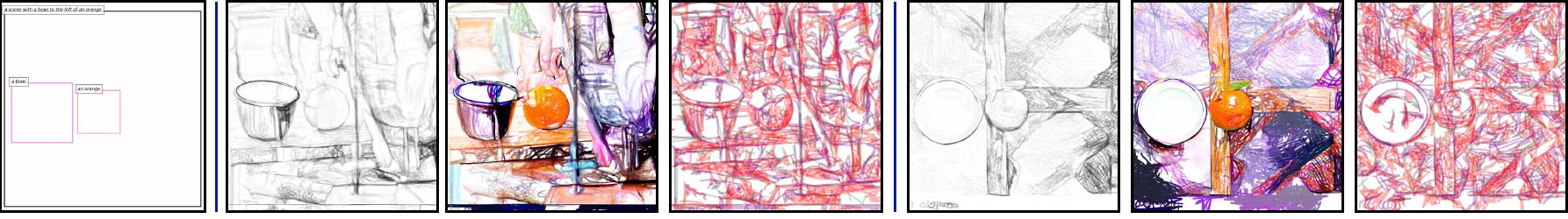}}\hfill \\
\subfloat[A \colorbox{yellow}{cup} above a \colorbox{yellow}{clock}]{\includegraphics[width=0.8\textwidth]{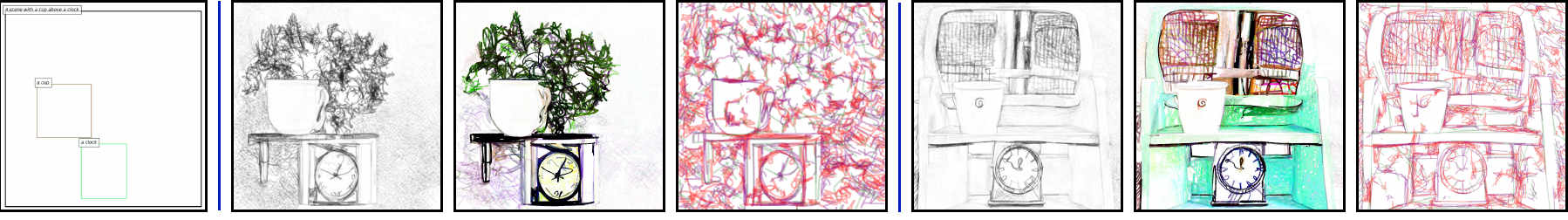}}\hfill \\
\subfloat[A \colorbox{yellow}{carrot} to the right of a \colorbox{yellow}{pizza}]{\includegraphics[width=0.8\textwidth]{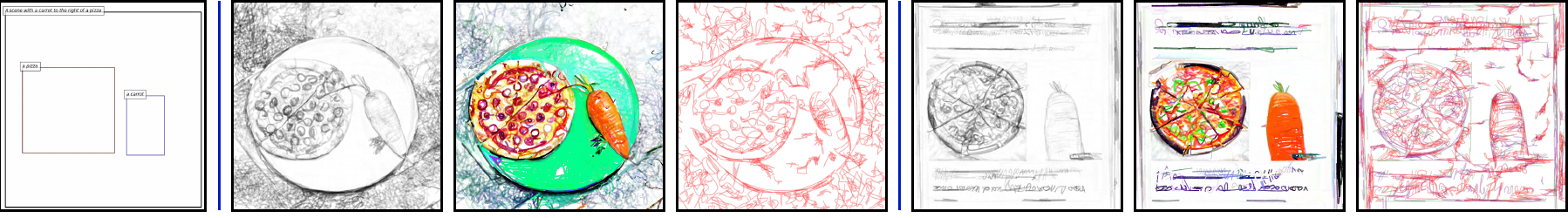}}\hfill \\
\subfloat[A \colorbox{yellow}{bowl} to the right of a \colorbox{yellow}{snowboard}
]{\includegraphics[width=0.8\textwidth]{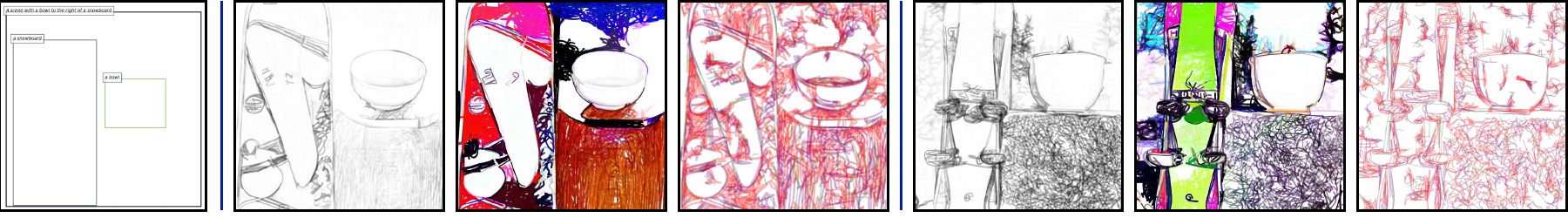}}\hfill \\
\subfloat[a \colorbox{yellow}{cat} standing next to an open \colorbox{yellow}{refrigerator} door]{\includegraphics[width=0.8\textwidth]{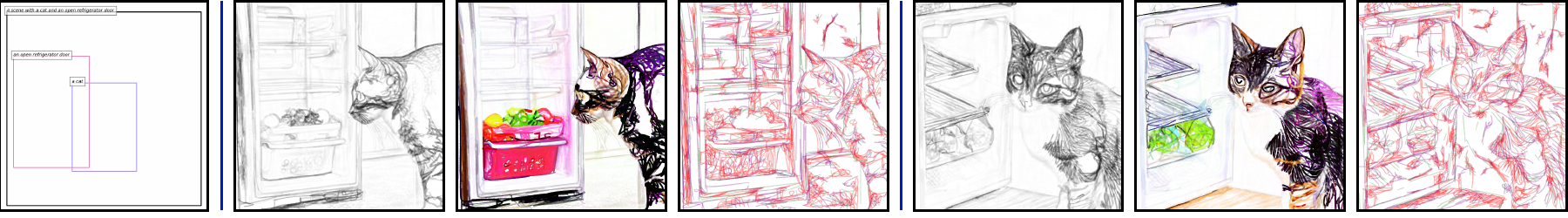}}\hfill \\
\subfloat[A \colorbox{yellow}{dog} to the left of a \colorbox{yellow}{cat}]{\includegraphics[width=0.8\textwidth]{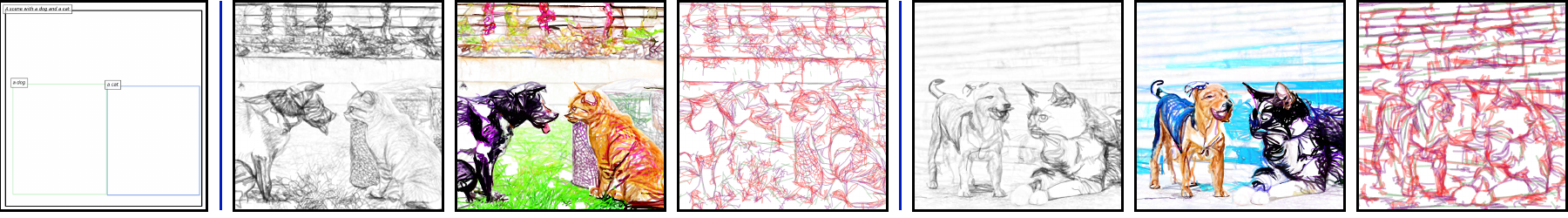}}\hfill \\
++\subfloat[A \colorbox{yellow}{bottle} of water is placed to the right of a \colorbox{yellow}{monitor}]{\includegraphics[width=0.8\textwidth]{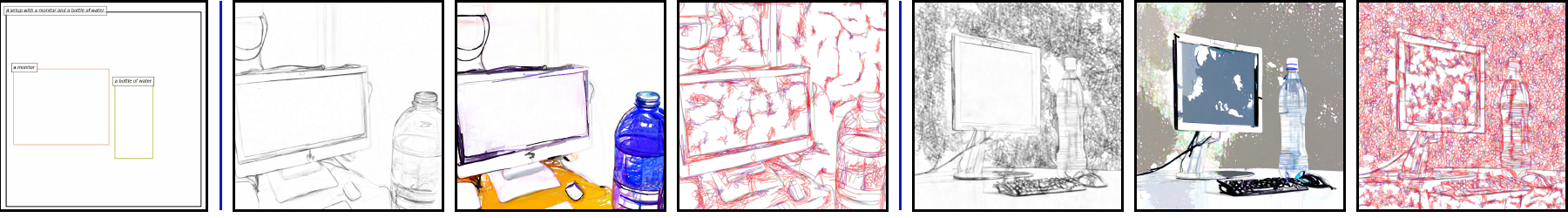}}\hfill
%\Description[]{}
\caption{\textbf{Spatial arrangement:} CraftSVG can easily determine the object position following the direction left, right, above, next and so on.}\label{fig:spatial}
%\vspace{-4mm}
\end{figure*}

\myparagraph{Spatial Arrangement:}
Here, we try to synthesize SVGs maintaining the spatial relationship (right, left, above, and so on). \cref{fig:spatial} demonstrates the performance of the CraftSVG with complex spatial arrangement prompts. This shows that CraftSVG has the potential to synthesize the SVGs following the arrangement described in the prompt.

\begin{figure*}[!ht]
\centering
\subfloat[A sketching with watercolors of a modern \colorbox{yellow}{Athens} neighborhood]{\includegraphics[width=0.95\textwidth]{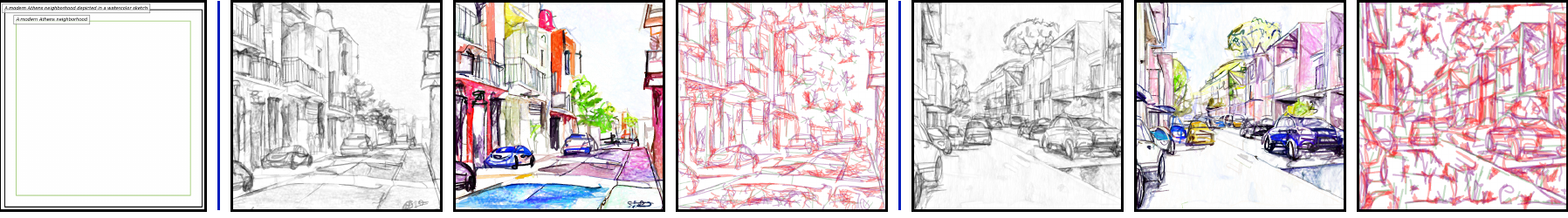}}\hfill \\
\subfloat[Real photo of Sydney \colorbox{yellow}{opera house}
]{\includegraphics[width=0.95\textwidth]{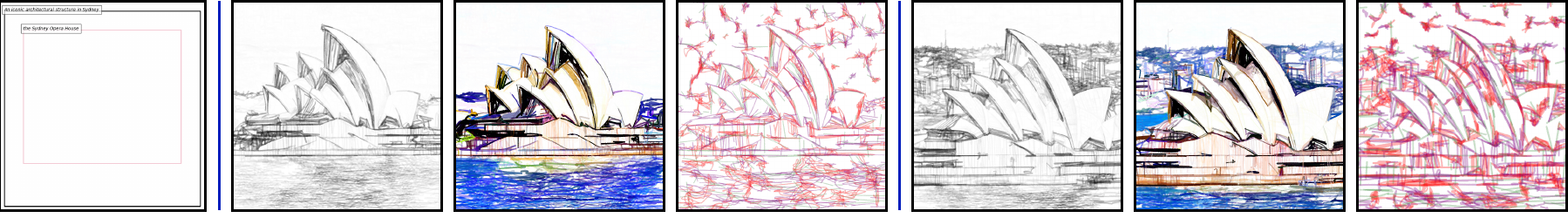}} \hfill \\
\subfloat[A detailed portrait of \colorbox{yellow}{Eiffel Tower}, incredibly detailed and realistic, 8k, sharp focus
]{\includegraphics[width=0.95\textwidth]{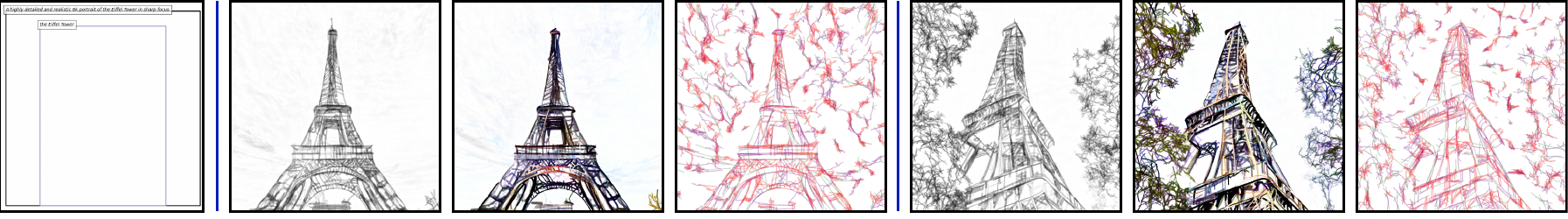}} \hfill \\
\subfloat[\colorbox{yellow}{Electronic board style buildings at New York city silhouette}
]{\includegraphics[width=0.95\textwidth]{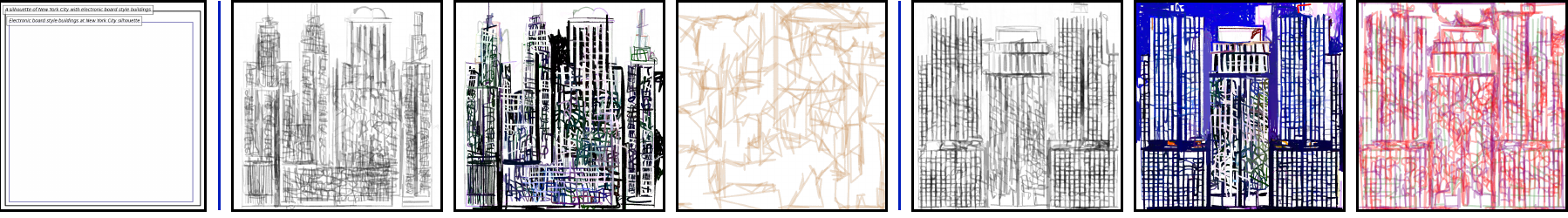}} \hfill \\
\subfloat[\colorbox{yellow}{Macaw} full color, ultra-detailed, realistic, insanely beautiful
]{\includegraphics[width=0.95\textwidth]{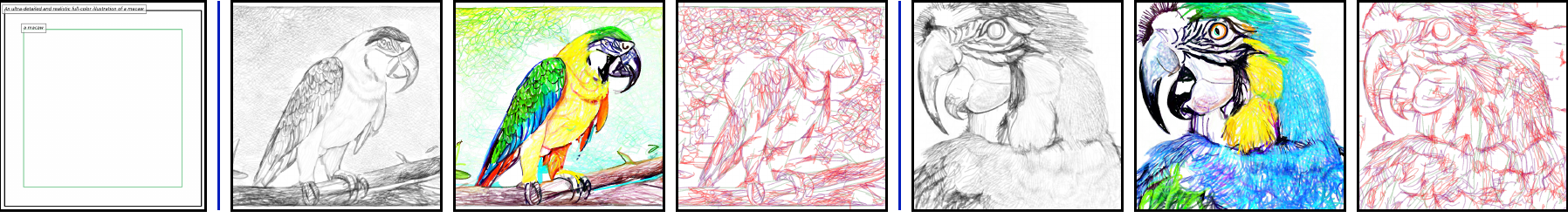}} \hfill \\
\subfloat[Potrait of \colorbox{yellow}{Latin woman} having a spiritual awakening, eyes closed, slight smile, illuminating lights, oil painting, by Van Gogh
]{\includegraphics[width=0.95\textwidth]{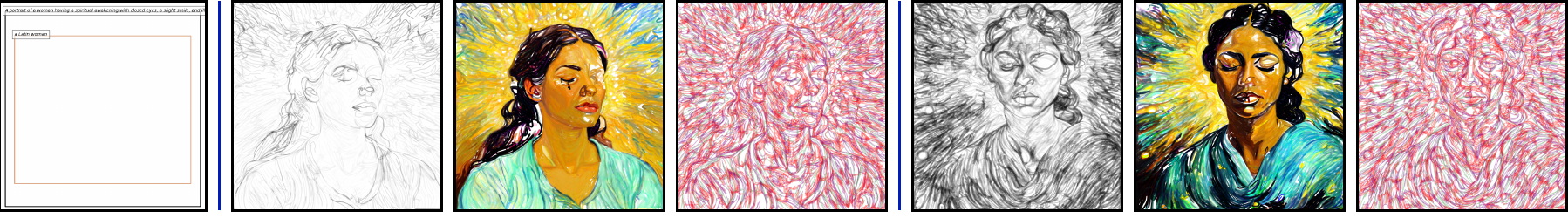}} \hfill \\
\subfloat[A tiny \colorbox{yellow}{analog clock}
]{\includegraphics[width=0.95\textwidth]{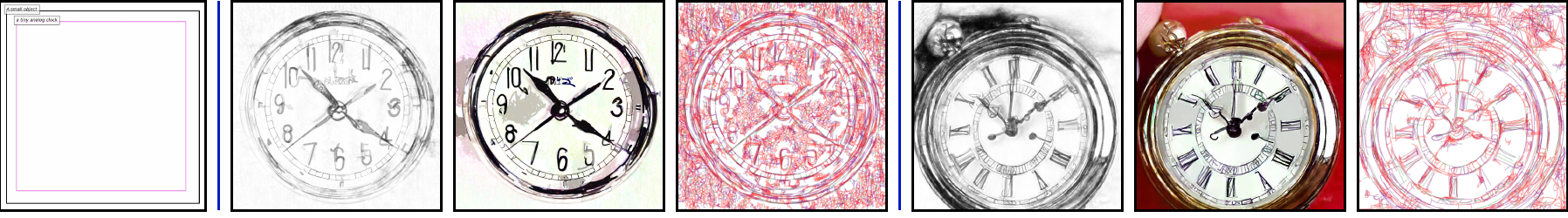}} \hfill
%\Description[]{}
\caption{\textbf{Object detailing:} CraftSVG produces detailing with user-specific style except for the human faces.}\label{fig:detail}
\end{figure*}

\myparagraph{Details of single object:}
Object detailing is one of the crucial components of human drawing as it is one of the significant ways to demonstrate the artist's capability to reproduce reality in detail. In order to replicate the same in synthesizing SVGs, we fed CraftSVG some prompts containing only single objects with specified details. As presented in \cref{fig:detail}, CraftSVG demonstrates the ability to create detailed and precise objects, with the exception of facial features. Investigating the latter aspect further could serve as a compelling avenue for future research.
% it can concluded that CraftSVG has the potential to synthesize objects with details and precision except the case of generating facial features. The latter could be an interesting future direction of this work.

\begin{figure*}[!ht]
\centering
\subfloat[Portrait of two white \colorbox{yellow}{bunnies}, super realistic, highly detailed]{\includegraphics[width=0.95\textwidth]{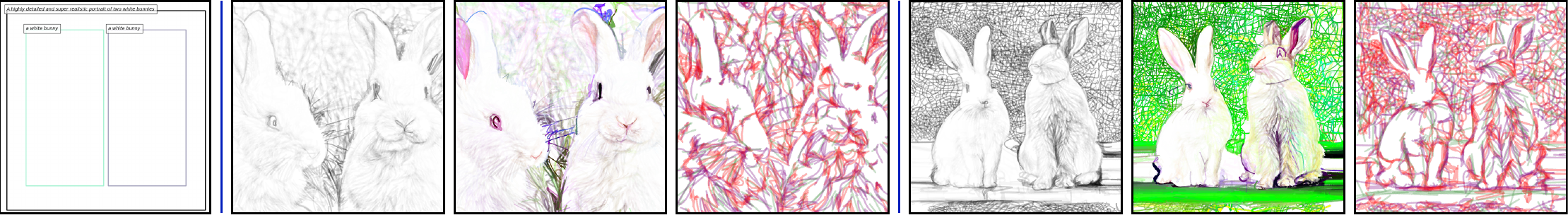}}\hfill \\
\subfloat[A \colorbox{yellow}{woman} is riding her \colorbox{yellow}{bike} down the street in front of traffic
]{\includegraphics[width=0.95\textwidth]{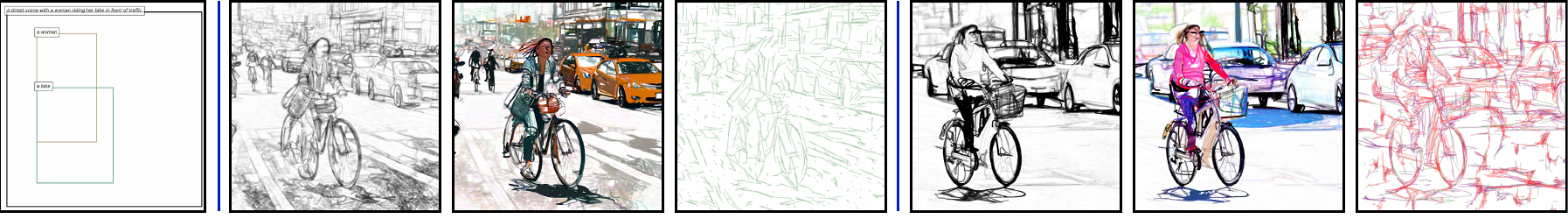}} \hfill \\
\subfloat[\colorbox{yellow}{Children} are sitting on the side of the \colorbox{yellow}{boat} in the water
]{\includegraphics[width=0.95\textwidth]{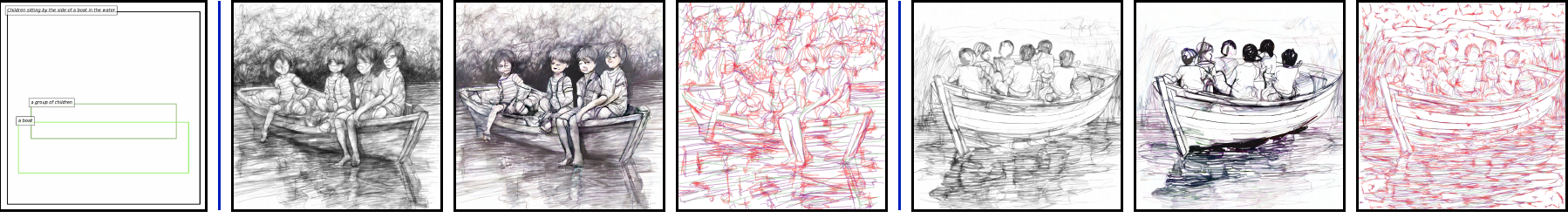}} \hfill \\
\subfloat[An angry \colorbox{yellow}{woman} staring at \colorbox{yellow}{coworkers}
]{\includegraphics[width=0.95\textwidth]{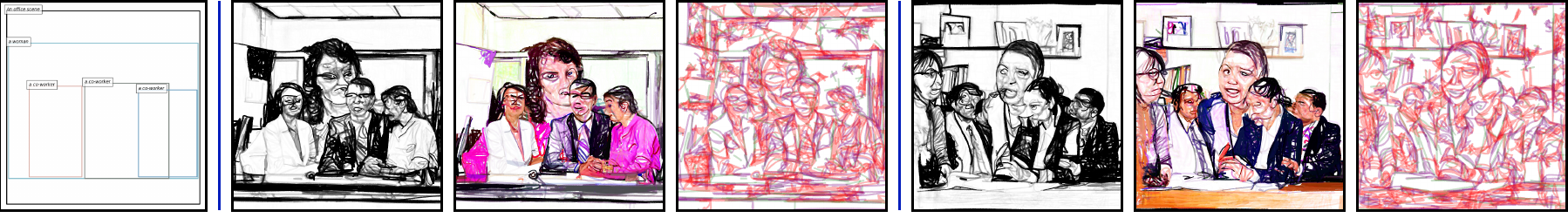}} \hfill \\
\subfloat[Three \colorbox{yellow}{books} and a \colorbox{yellow}{laptop} kept neatly in a workstation
]{\includegraphics[width=0.95\textwidth]{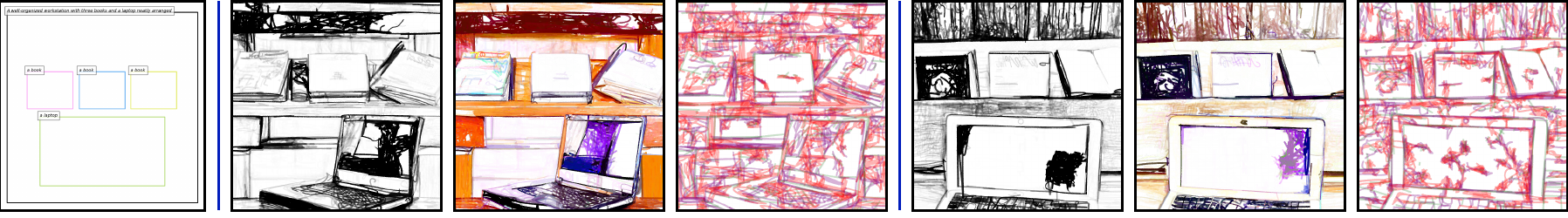}} \hfill \\
\subfloat[A pair of \colorbox{yellow}{scissors} and a \colorbox{yellow}{knife} resting in a \colorbox{yellow}{desk organizer}
]{\includegraphics[width=0.95\textwidth]{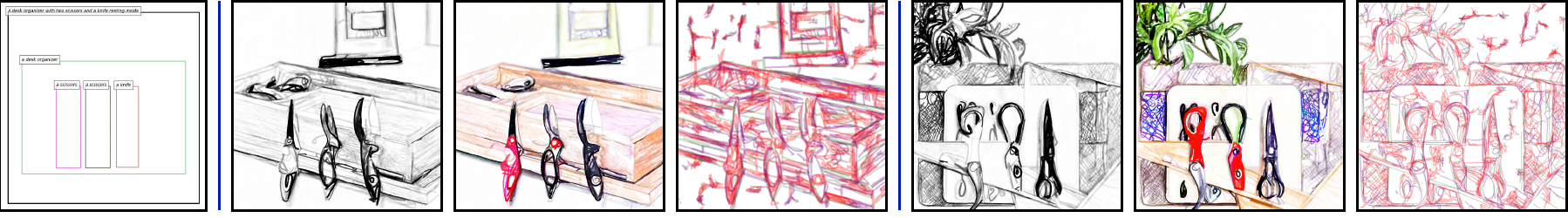}} \hfill \\
\subfloat[A \colorbox{yellow}{laptop} computer a \colorbox{yellow}{keyboard} and two \colorbox{yellow}{monitors}
]{\includegraphics[width=0.95\textwidth]{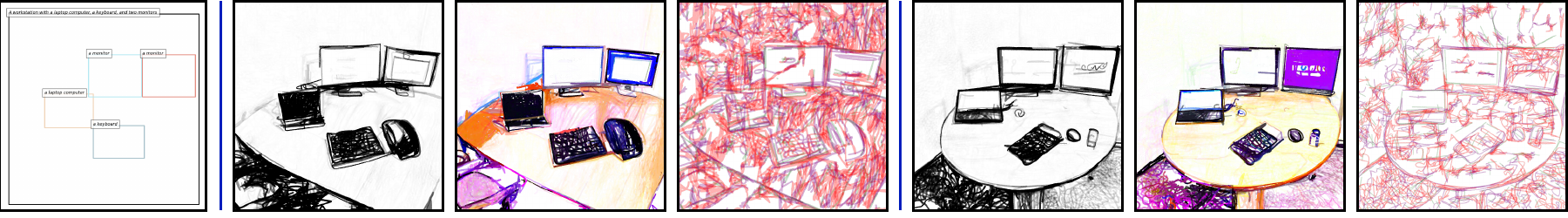}} \hfill
%\Description[]{}
\caption{\textbf{Miscellaneous:} with different types of descriptions CraftSVG successfully produces the desired SVGs specified in the text prompt.}\label{fig:misc}
\end{figure*}

\myparagraph{Miscellaneous}
Here we mixed all types of prompts with enumeration, spatial relationship, detailing, behavioral expression, and so on. From \cref{fig:misc}, it can be concluded that CraftSVG successfully tackles all the complexity provided in the text prompt.
